\documentclass[10pt,twocolumn,letterpaper]{article}

\usepackage{cvpr}              % To produce the CAMERA-READY version
\usepackage{amsthm} % 加载数学定理包
\usepackage{multirow}
\usepackage{booktabs}
\usepackage{lineno}
\usepackage[T1]{fontenc}
\usepackage{algorithm}
\usepackage{algorithmic}
\newcommand\blfootnote[1]{%
\begingroup
\renewcommand\thefootnote{}\footnote{#1}%
\addtocounter{footnote}{-1}%
\endgroup
}
\definecolor{cvprblue}{rgb}{0.21,0.49,0.74}
\usepackage[pagebackref,breaklinks,colorlinks,allcolors=cvprblue]{hyperref}

\def\paperID{10687} % *** Enter the Paper ID here
\def\confName{CVPR}
\def\confYear{2025}

\title{Adaptive Unimodal Regulation \\for Balanced Multimodal Information Acquisition}

\author{Chengxiang Huang\textsuperscript{1,$\dagger$} \hspace{1em} Yake Wei\textsuperscript{2,$\dagger$} \hspace{1em} Zequn Yang\textsuperscript{2} \hspace{1em} Di Hu\textsuperscript{2,*}\\
\textsuperscript{1}Beijing University of Posts and Telecommunications \hspace{2em} \textsuperscript{2}Renmin University of China\\
{\tt\small huangchengxiang2021@bupt.edu.cn, \{yakewei,zqyang,dihu\}@ruc.edu.cn}
}

\begin{document}
\maketitle
% 使用标准脚注命令
\begin{abstract}
Sensory training during the early ages is vital for human development. Inspired by this cognitive phenomenon, we observe that the early training stage is also important for the multimodal learning process, where dataset information is rapidly acquired. We refer to this stage as the prime learning window. However, based on our observation, this prime learning window in multimodal learning is often dominated by information-sufficient modalities, which in turn suppresses the information acquisition of information-insufficient modalities.
To address this issue, we propose \textbf{Info}rmation Acquisition \textbf{Reg}ulation (InfoReg), a method designed to balance information acquisition among modalities. Specifically, InfoReg slows down the information acquisition process of information-sufficient modalities during the prime learning window, which could promote information acquisition of information-insufficient modalities. This regulation enables a more balanced learning process and improves the overall performance of the multimodal network. Experiments show that InfoReg outperforms related multimodal imbalanced methods across various datasets, achieving superior model performance. The code is available at \url{https://github.com/GeWu-Lab/InfoReg_CVPR2025}.
\blfootnote{\noindent
\textsuperscript{$\dagger$}Equal contribution. 
\textsuperscript{*}Corresponding author.
}
\end{abstract}    

\section{Introduction}
\label{sec:intro}
% Just as early learning experiences shape life-long abilities in humans and animals, the absence of specific training during early developmental stages can result in sensory deficits that lead to persistent skill impairments that later interventions cannot fully correct \cite{kandel2000principles,hensch2004critical}. For instance, the absence of specific training during early developmental stages, such as depriving individuals of visual input \cite{wiesel1963single,wiesel1982postnatal} or auditory input \cite{arIARga2011mice}, can cause sensory deficits.

Learning during early developmental ages in humans and animals is important for skill impairments \cite{kandel2000principles,hensch2004critical,wiesel1982postnatal}. Similarly, in deep learning, recent studies have found that models learn in stages, with early learning being especially important for effective information acquisition \cite{achille2018critical,jastrzkebski2018relation}.

\begin{figure}[t]
  \centering
  \begin{subfigure}[t]{0.49\linewidth} % 这里使用 0.48\linewidth 以实现并排效果
    \centering
    \includegraphics[width=\linewidth]{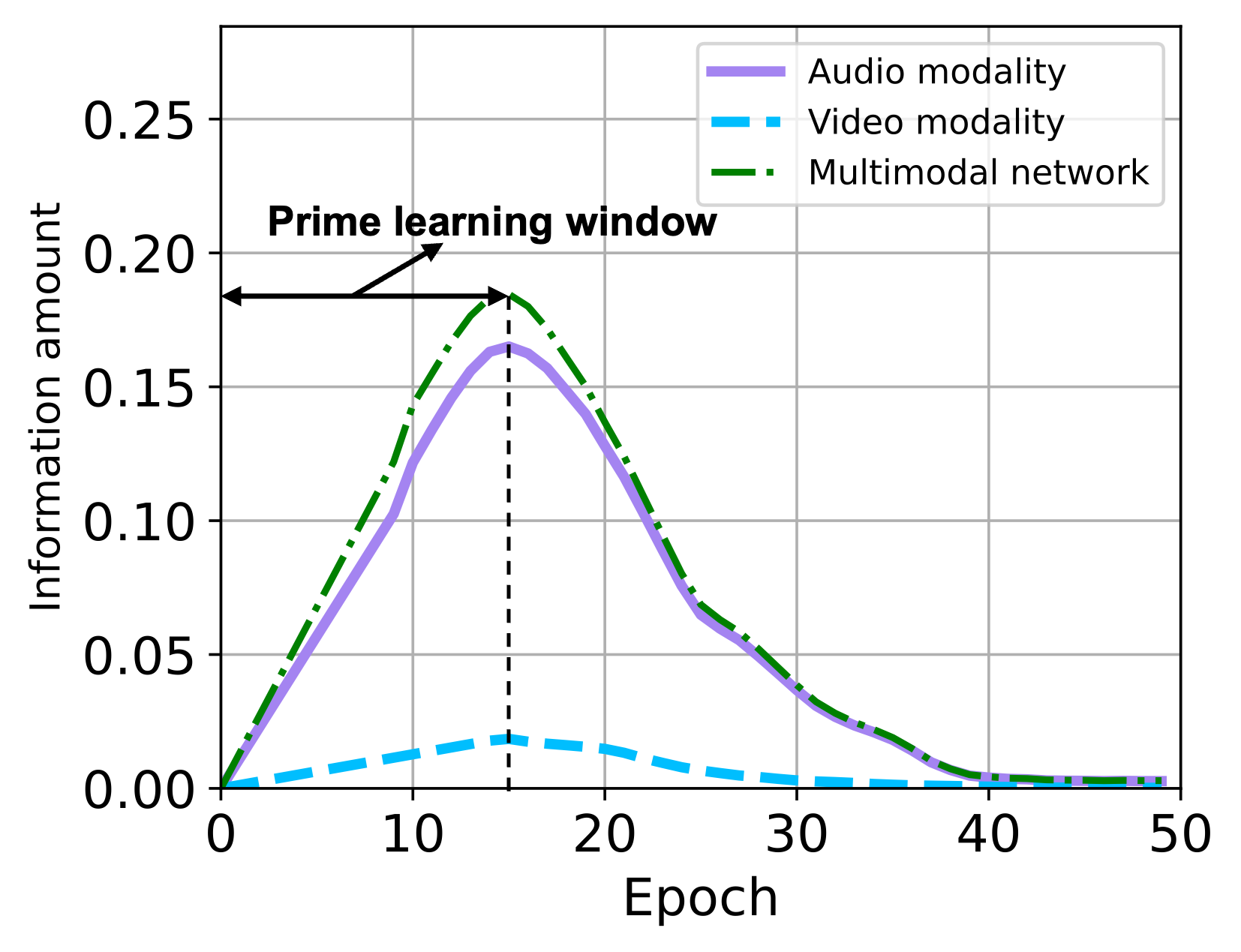}
    \caption{Multimodal scenario.}
    \label{fig:mmclp}
  \end{subfigure}
  \hfill % 用于增加子图之间的间距
  \begin{subfigure}[t]{0.49\linewidth} % 这里使用 0.48\linewidth 以实现并排效果
    \centering
    \includegraphics[width=\linewidth]{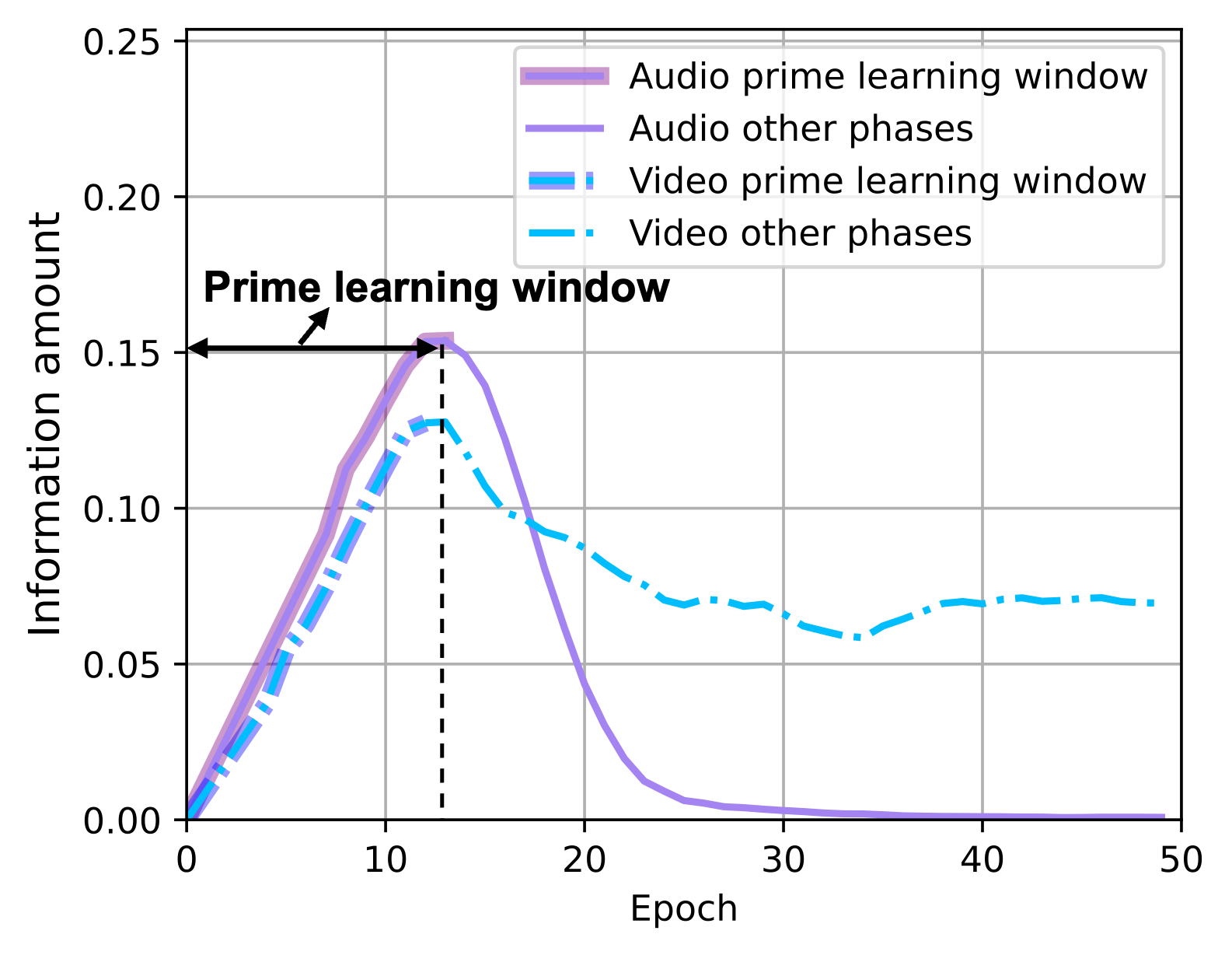}
    \caption{Unimodal scenario.}
    \label{fig:uniclp}
  \end{subfigure}%
  \caption{\textbf{(a).} Information amount variation of the audio encoder, video encoder, and multimodal model during the training process on CREMA-D~\cite{cao2014crema}. \textbf{(b).}  Information amount variation of the audio and video modalities when trained independently on CREMA-D.  }
  \label{fig:CLP}
\vspace{-15pt}
\end{figure}

The above circumstance motivates us to investigate the process of information acquisition in the multimodal scenario. Due to the presence of multiple modalities, multimodal learning networks are expected to capture sufficient information from each modality \cite{gao2020listen,wang2021temporal}. To observe the information acquisition of different modalities, we use the trace of the Fisher Information Matrix \cite{achille2018critical,fisher1925theory} to investigate the information amount of each modality in multimodal models.
However, based on our observation, the amount of information in multimodal models is not well consistent with intuitive expectations.
\textit{Firstly}, as shown in Figure \ref{fig:mmclp}, the green curve, representing the overall multimodal network, demonstrates a rapid increase in information amount during the early periods. Information acquisition is most rapid in this early stage, which we refer to as \textit{ the prime learning window}, reaching a peak at the end of this period, followed by a subsequent decline.
\textit{Secondly}, for audio modality, its overall trend closely aligns with the overall multimodal model, and shows a high information acquisition amount during the prime learning window. These findings align with our expectation that modality information could be acquired effectively within the prime learning window.
However, the video modality, represented by the blue curve, shows a much lower information acquisition amount during the prime learning window. Although the video modality is capable of effective information acquisition in the unimodal scenario in Figure \ref{fig:uniclp}, it fails to do so in the multimodal scenario when trained jointly with the audio modality. 
These observations suggest that information-insufficient modalities, like video, experience suppressed information acquisition during the prime learning window due to the stronger information acquisition capacity of information-sufficient modalities, such as audio. Moreover, as the multimodal model’s capacity for information acquisition diminishes in later stages, the imbalance observed during the prime learning window cannot be compensated for by simply extending the training time, as demonstrated in Table \ref{tab:loss_metrics}.

\begin{table}[t]
\centering
\begin{tabular}{l|c}
\bottomrule
\textbf{Method} & \textbf{Accuracy} \\
\hline
Video in Joint training (50 epochs)  & 35.14 \\
Video in Joint training (100 epochs) & 35.36 \\
Video in InfoReg (50 epochs) & \textbf{49.65} \\
\toprule
\end{tabular}
\vspace{-5pt}
\caption{ Comparison of performance for the video modality between Joint training and InfoReg on the CREMA-D dataset. Simply extending the training time (100 epochs) cannot compensate for the suppressed video modality. }
\label{tab:loss_metrics}
\vspace{-20pt}
\end{table}

Recent studies have investigated the problem of imbalanced learning across modalities in multimodal learning \cite{wang2020makes,peng2022balanced,ismail2020improving}. Although several studies have made progress in addressing this issue \cite{du2021improving,sun2021learning,weimmpareto,li2023boosting},  their methods typically apply adjustments across the entire training process without recognizing the importance of the prime learning window. Consequently, the effectiveness of these methods is significantly constrained. Our research reveals that the prime learning window plays a vital role in multimodal learning. In this window, there is a significant imbalance in information acquisition across modalities. Therefore, effective adjustment within this window is essential for achieving balanced information acquisition across modalities.

Based on the above analysis, information-insufficient modalities experience a significantly reduced information acquisition amount during the prime learning window due to suppression from information-sufficient modalities. To address this imbalance, we slow the information acquisition rate of information-sufficient modalities within this window, thereby allowing information-insufficient modalities to improve their acquisition. Hence, we propose our method, \textbf{Info}rmation Acquisition \textbf{Reg}ulation (InfoReg). In InfoReg, the process begins by determining whether information-sufficient modalities are within the prime learning window. If so, a unimodal regulation term is applied to regulate the Fisher Information \cite{achille2018critical,fisher1925theory}, thereby restricting these modalities acquire information. As a result, InfoReg promotes a more balanced information acquisition across modalities, enhancing the overall performance of the model. Our extensive experiments across multiple datasets show that InfoReg achieves superior performance and improves modality balance by helping information-insufficient modalities acquire more information during the prime learning window. Additionally, InfoReg enhances feature quality, supporting prior research on the link between early-stage information acquisition and robust feature representation \cite{achille2018emergence,achille2018critical}. These results collectively validate the effectiveness of our approach.

Our main contributions are summarized as follows:

\begin{itemize}

    \item \textbf{Firstly}, we identify the prime learning window in multimodal learning, a critical period where imbalances in information acquisition significantly impact modality balance and overall performance.
    
    \item \textbf{Secondly}, We analyze the imbalance of Fisher Information among modalities and propose InfoReg, a method that regulates the information acquisition of information-sufficient modalities during the prime learning window.
    
    \item \textbf{Finally}, we validate InfoReg on multiple datasets and settings, demonstrating its considerable improvement while maintaining balanced performance across modalities.

\end{itemize}

\section{Related Work}
\label{sec:related work}
\subsection{Multimodal imbalance learning}
Jointly training multiple modalities is intuitively expected to enhance performance \cite{liang2022foundations}. However, imbalanced learning across modalities poses substantial challenges, hindering the effective training of multimodal networks \cite{huang2022modality,team2024chameleon,aghajanyan2023scaling,the2024large}.
Previous studies have investigated the imbalanced learning problem across modalities in multimodal learning networks \cite{wang2020makes,peng2022balanced}, where certain modalities tend to dominate, limiting the learning of other modalities. This imbalance can degrade overall performance and lead to significant disparities between modalities, sometimes even causing multimodal learning to underperform compared to single-modality scenarios \cite{peng2022balanced}. To address this problem, many methods \cite{wang2020makes, peng2022balanced, li2023boosting, weimmpareto, wei2024diagnosing} have been proposed, primarily focusing on balancing the optimization of each modality throughout the training process.  BalanceBench \cite{xu2025balancebenchmark} further categorizes these methods based on their distinct characteristics, offering a comprehensive framework to evaluate their effectiveness and limitations. Specifically, OGM \cite{peng2022balanced} balances modalities by adjusting the gradients of well-learned modalities. MMPareto \cite{weimmpareto} leverages an optimized Pareto front to balance the performance across modalities, aiming to improve the generalization of multimodal models. 
Despite the successes of previous work, they have largely overlooked the impact of different training periods on information acquisition across various modalities. We have recognized this issue and designed a method accordingly. By slowing down the information acquisition rate of information-sufficient modalities, our approach alleviates the suppression of information-insufficient modalities, allowing them to acquire more information in the early stages, effectively balancing the performance of different modalities. This leads to considerable improvements in both overall performance and modality balance.

\subsection{Early stages in deep learning}
The learning process of deep learning models consists of two main phases: an initial phase of information acquisition from the dataset, followed by a phase of gradual information compression or forgetting \cite{jastrzebski2021catastrophic,jastrzkebski2018relation,achille2018emergence,yan2021critical}. Recent studies highlight the crucial role of early-stage learning in shaping feature representation and overall model performance \cite{frankle2020early,golatkar2019time,jastrzebski2020break,lewkowycz2020large,yan2023criticalfl}, similar to findings in neuroscience \cite{kandel2000principles,hensch2004critical,wiesel1963single,arriaga2011mice}. Further, the information missed during the early stages cannot be recovered by extending the training duration later on \cite{achille2018critical}. 
We define this early important learning period as the prime learning window. However, previous research has overlooked the importance of the prime learning window in multimodal learning. Our findings reveal that balancing information acquisition among modalities during this window leads to considerable performance improvements in multimodal models.

\section{Method}

\subsection{Multimodal learning framework}
For convenience, let the dataset be denoted by \( D = \{(x_n, y_n)\}_{n=0,1,\dots, N-1} \), where each \( x_n \) contains inputs from \( M \) modalities: \( x_n = (x_n^1, x_n^2, \dots, x_n^M) \). The target label \( y_n \in \{1, 2, \dots\} \) represents the class of sample \( x_n \). For each modality \( m \), where \( m \in \{1, 2, \dots, M\} \), the input is processed through the corresponding encoder \( \varphi^m(w_m, \cdot) \). Here, \( w_m \) are the weights of encoder \( m \). After feature extraction, the outputs are concatenated and passed to a single-layer linear classifier. Finally, one joint multimodal cross-entropy loss $\mathcal{L}_{joint}(  {w})$ is utilized to optimize the model.

% Let \( W \in \mathbb{R}^{C \times \sum_{m=1}^{M} d_{\varphi^m}} \) and \( b \in \mathbb{R}^C \) denote the parameters of the linear classifier, where \( d_{\varphi^m} \) is the output dimension of \( \varphi^m(\theta_m, \cdot) \).
% To optimize the model across all modalities, we employ a multi-task multimodal loss function.
%  The overall loss function can be defined as follows:
% \begin{equation}
% \label{equ:mmloss}
% \begin{split}
%     &\min _{w} \mathcal{L} := \mathcal{L}_{joint}(  {w})+ \sum_{i = 1}^{M} \mathcal{L}_{i}({w_i}),  
% \end{split}
% \end{equation} where $\mathcal{L}_{joint}$ denotes the loss associated with joint learning across all modalities, $\mathcal{L}_{i}$ represents the unimodal loss for each individual modality, and $w$ corresponds to the weights assigned to different modalities. This loss function is designed to incorporate unimodal losses to enhance the overall learning capability of the multimodal system. We assume that all loss components are cross-entropy losses.

\begin{figure*}[ht]
\centering
\includegraphics[width=0.75\textwidth]{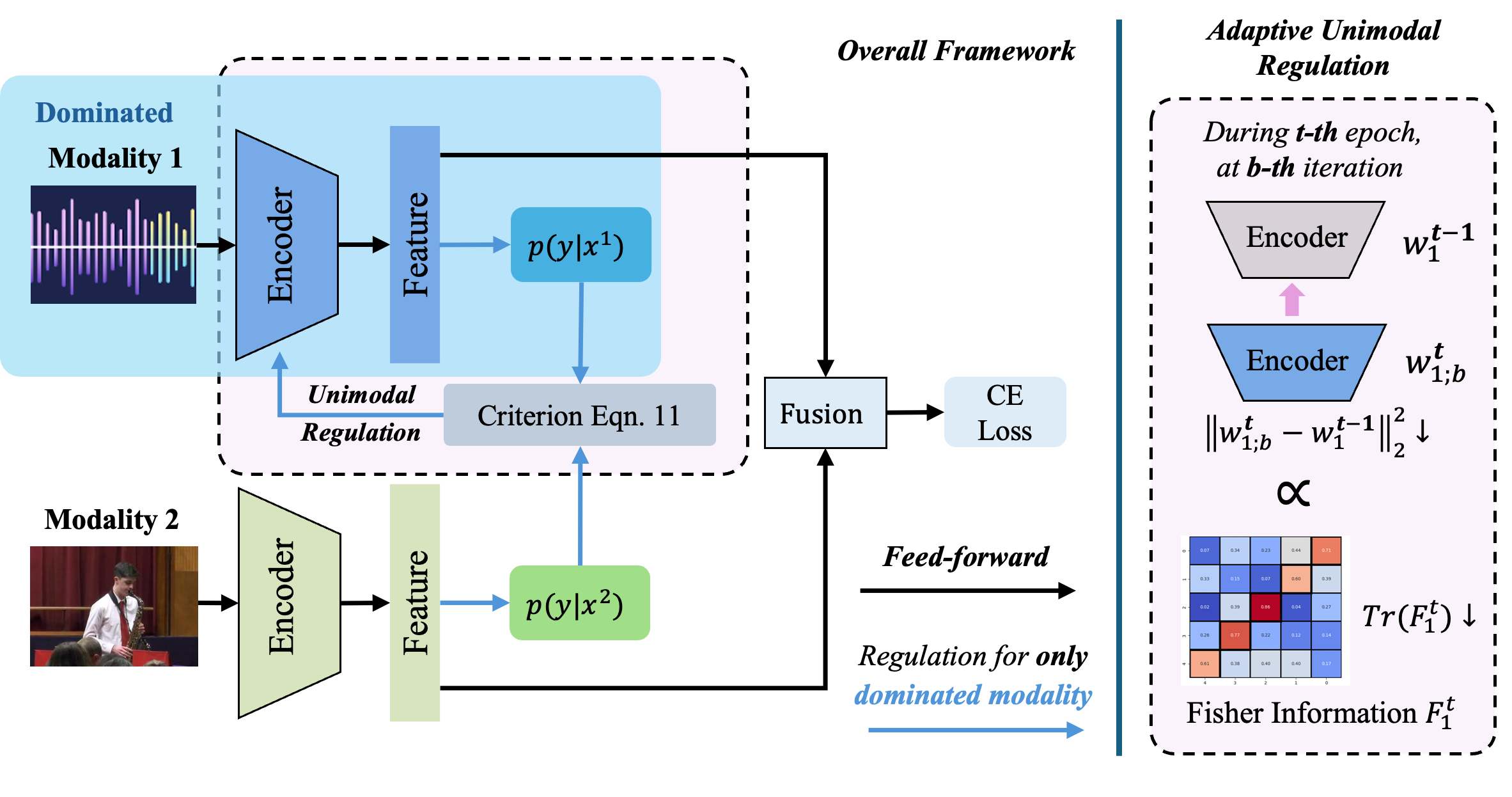} % Reduce the figure size so that it is slightly narrower than the column. Don't use precise values for figure width.This setup will avoid overfull boxes. 
\vspace{-10pt}
\caption{\textbf{Overview of InfoReg.} This figure shows the main components and workflow of InfoReg. The left side presents our overall framework, while the right side highlights the adaptive unimodal regulation. During the training, InfoReg first identifies the information-sufficient modalities, then evaluates whether they are in the prime learning window, and finally applies adaptive unimodal regulation. }
\label{fig:overview}
\vspace{-10pt}
\end{figure*}

\subsection{Fisher Information in multimodal learning}
Following previous work \cite{achille2018emergence,shwartz2017opening} that uses the Kullback-Leibler (KL) divergence to measure the information contained in the weights of a unimodal network, the information acquisition process for a single modality $m$ can be evaluated by this metric. Given the posterior distribution $p_{w_m}(y|x;D)$, encoded by the unimodal encoder with weights $w_m$ and its prior distribution $q_{w_m}(y|x)$, the mutual information is defined as follows:
\begin{equation}
\label{equ:divergence}
D_{KL}\left ( p_{w_m} \parallel q_{w_m} \right ) = \int p_{w_m}(y|x;D)log\frac{p_{w_m}(y|x;D)}{q_{w_m}(y|x)}dy.
\end{equation}
However, quantifying the information a unimodal encoder acquires from the dataset is challenging because the ground truth prior distribution $q_{w_m}(y|x)$ is not accessible.
As an alternative, inspired by \cite{achille2018critical}, the rate of information acquisition from the dataset can be estimated by calculating the KL divergence between distributions encoded by weights at successive moments in training. Specifically, given a perturbation $w_m^{'} = w_m + \delta w_m$. The discrepancy between the distributions $p_{w_m}(y|x;D)$ and $p_{w^{'}_m}(y|x;D)$ reflects the rate of information acquisition and can be defined as:
\begin{equation}      
\label{equ:divergence2}
D_{KL}(p_{w_{m}}\parallel p_{w_m^{'}})=\int p_{w_m}(y|x;D)log\frac{p_{w_m}(y|x;D)}{p_{w_m^{'}}(y|x;D)}dy. 
\end{equation}
Further, the KL divergence can be approximated to second order by applying the Taylor expansion:
\begin{equation}
\label{equ:divergence3}
D_{KL}(p_{w_{m}}\parallel p_{w_m}^{'}) \approx \frac{1}{2} \delta w_m^TF_m \delta w_m,
\end{equation} where $F_m$ is the Fisher Information Matrix (FIM) \cite{fisher1925theory}, and is  defined as:
\begin{equation}
\label{equ:FIM}
    F_m = \mathbb{E}_{y \sim p_w} \left[ \nabla _{w_m}logp_{w_m}(y|x)\nabla _{w_m}logp_{w_m}(y|x)^T\right].
\end{equation}
The FIM plays a crucial role in quantifying the amount of information captured by the deep neural network \cite{kirkpatrick2017overcoming,achille2018critical} and acts as a local measure, assessing how small perturbations in the model’s parameters influence its output \cite{amari2000methods}. Additionally, the FIM is a semi-definite approximation of the Hessian matrix, providing insights into the curvature of the loss landscape at a given point during training \cite{thomas2020interplay,martens2020new}. 

In multimodal learning, for a unimodal encoder $\varphi ^{m}$, the gradient of the encoder can be expressed as :
\begin{equation}
\label{equ:gradient}
    g_{\varphi^{m}}(w_m,x_n^m) = \nabla _{w_m}logp_{w_m}(y_n|x^{m}_n) = \nabla _{w_m}\mathcal{L}_{joint}(  {w_m}).
\end{equation}
Based on this, the Fisher Information Matrix $F_m$ can be reformulated as:
\begin{equation}
\label{equ:FIM2}
    F_m = \mathbb{E}_{x_n^{m} \sim X^{m}} 
    \left[ g_{\varphi^{m}} \left( w_m, x^{m}_n \right) 
    g_{\varphi^{m}} \left( w_m, x^{m}_n \right)^T \right ].
\end{equation}
However, computing $F_m$ directly is computationally expensive. To address this, we use the trace of the Fisher Information Matrix, denoted as $Tr(F_m)$, to measure the amount of information captured by the deep neural network. This trace can be computed more efficiently and is defined as:
\begin{equation}
\label{equ:TrF}
    Tr(F_m) =\mathbb{E}_{x_n^{m} \sim X^{m}} \left[ \parallel g_{\varphi^{m}} \left( w_m, x^{m}_n \right) \parallel ^2
     \right ].
\end{equation}
As shown in Figure \ref{fig:CLP}, $Tr(F_m)$ could effectively measure the amount of information acquired and identify the prime learning window.

% For simplicity, we denote $g_{\varphi^{m}}(w_m,x_n^m)$ as $g_{\varphi^{m}}$. According to Equation \ref{equ:TrF}, it follows that 
% \begin{equation}
%     Tr(F_m) \sim \parallel g_{\varphi^{m}} \parallel ^2.
% \label{equ:Tr_g}
% \end{equation} 
As illustrated in Figure \ref{fig:gradient_diff}, information-sufficient modalities will exhibit significantly larger values of  $g_{\varphi^{m}}$ during the prime learning window. Due to the squared term in Equation \ref{equ:TrF}, these substantial differences in $g_{\varphi^{m}} $ between modalities are further amplified, thereby making the imbalance in Fisher Information even more pronounced. This indicates that information-sufficient modalities have a clear advantage in the information acquisition during the prime learning window and dominate the overall information acquisition of the multimodal model.

% As illustrated in the Figure \ref{fig:gradient_diff}, in multimodal learning models, some modalities tend to exhibit significantly larger values of  $g_{joint}^{m_i}$ during the prime learning window. These modalities are thus able to acquire more comprehensive information during this important learning phase. 

% As shown in Figure \ref{fig:trf_diff}, given the squared term, these already substantial differences in $g_{joint}^{m_i}$ between modalities are  further amplified, making disparities in information acquisition across modalities even more pronounced. This finding highlights that, in contrast to conventional deep learning, multimodal learning scenarios necessitate greater attention to imbalances in information acquisition during the prime learning window, especially given the influence of the multimodal loss.

\subsection{Information acquisition regulation}

To address the Fisher Information imbalance observed during the prime learning window, our method regulates the Fisher Information of information-sufficient modalities in this important period to slow down their information acquisition, thereby promoting information acquisition in other modalities.
Our method consists of the following main components:
\begin{figure}[t]
  \centering
  \begin{subfigure}[t]{0.50\linewidth} % 这里使用 0.48\linewidth 以实现并排效果
    \centering
    \includegraphics[width=\linewidth]{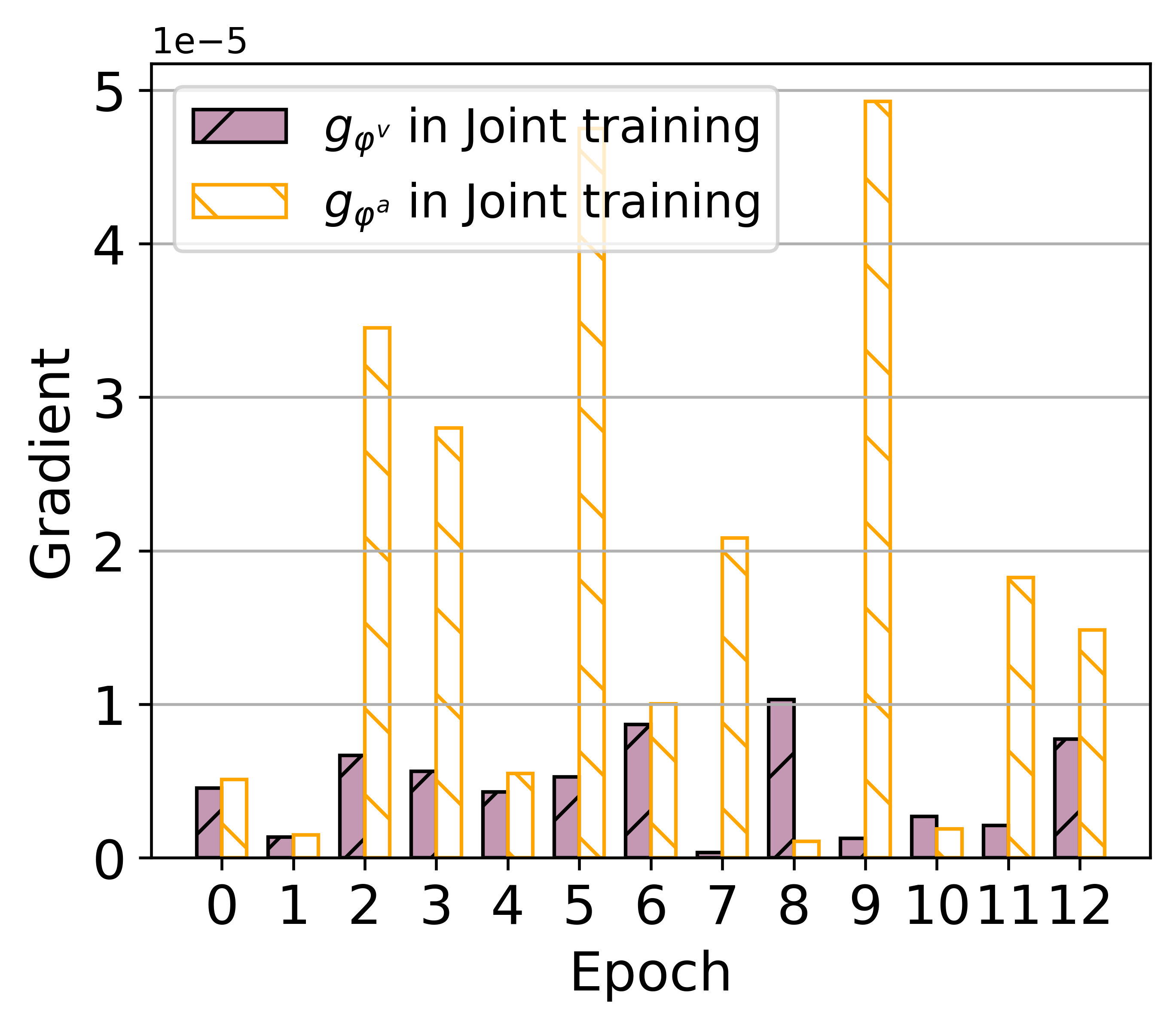}
    \caption{Gap between $g_{\varphi^m}$.}
    \label{fig:gradient_diff}
  \end{subfigure}%
  \hfill % 用于增加子图之间的间距
  \begin{subfigure}[t]{0.50\linewidth} % 这里使用 0.48\linewidth 以实现并排效果
    \centering
    \includegraphics[width=\linewidth]{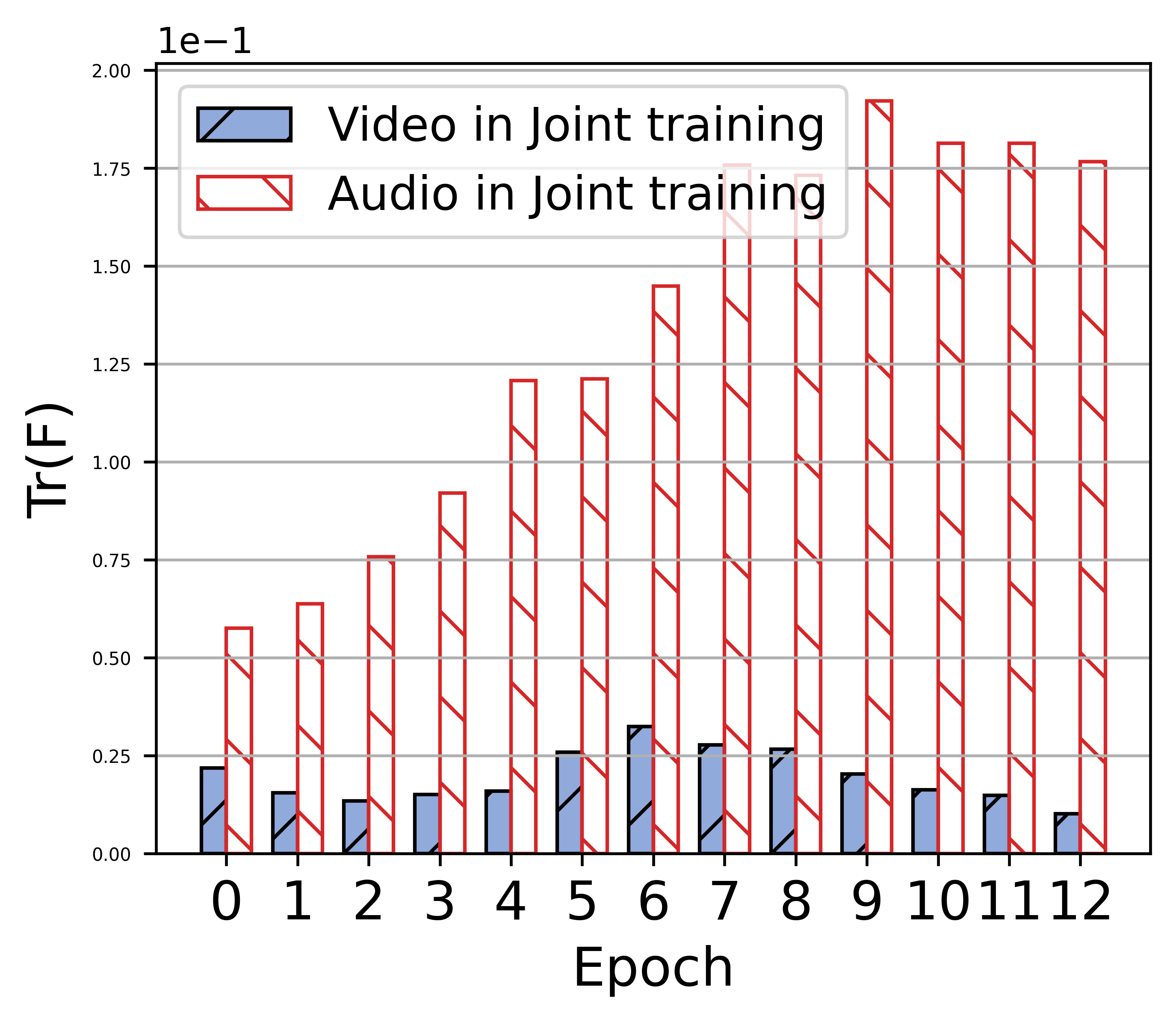}
    \caption{Gap between $Tr(F_m)$.}
    \label{fig:trf_diff}
  \end{subfigure}
  \caption{\textbf{(a).} The gradient gap between the audio modality and video modality on CREMA-D. \textbf{(b).} The $Tr(F_m)$ gap between the audio modality and video modality on CREMA-D.  }
  \label{fig:diff}
\vspace{-15pt} % 控制图题和下方文字之间的间距
\end{figure}

\begin{itemize}
    \item \textit{Evaluate the prime learning window for information-sufficient modalities}: We first identify the information-sufficient modalities and then evaluate whether they fall within the prime learning window.
    
    \item \textit{Adaptive unimodal regulation}: For information-sufficient modalities, we apply adaptive unimodal regulation to approximately regulate the Fisher Information.
\vspace{-5pt} % 控制图题和下方文字之间的间距
\end{itemize}

\subsubsection{Evaluating the prime learning window.}
Inspired by OGM \cite{peng2022balanced}, performance scores are used to identify information-sufficient modalities. Afterward, an assessment is made to determine whether these information-sufficient modalities are in the prime learning window. For each iteration, assume that training has reached epoch $t$ and is currently processing batch $b$, where $b \in \{0,1,\dots,B-1 \}$ and $B$ represent the total number of batches.
The performance score for each modality is given by:
\begin{equation}
\label{equ:score_j}
     s_{m;b}^t = \mathbb{E}_{x_n^{m} \sim X^{m}}\left [ - log\left ( softmax(\varphi^{m}\left ( x_n^{m} \right ) )_{y_n} \right ) \right ].
\end{equation}
Then, we define its performance gap $\Delta_m$ relative to other modalities to determine information-sufficient modalities during the prime learning window. Let $C_m$ represent the number of modalities with performance scores less than $s_{m;b}^t$:
\begin{equation}
    C_m = \left |  \left \{ m^{'}\in \left [ M \right ] \backslash \left \{ m \right \}  ;s_{m^{'};b}^{t} < s_{m;b}^{t} \right \} \right |.
\end{equation} The performance gap $\Delta_m$ can then be expressed as:
\begin{equation} 
\label{equ:delta}
\Delta_m = \frac{1}{C_m} \sum_{m^{'} \in [M] \backslash \{m\} ; s_{m^{'};b}^{t} < s_{m;b}^{t}} \left( s_{m;b}^{t} - s_{m^{'};b}^t \right), \end{equation} where $\Delta_m$ measures the average performance difference between $m$ and all lower or equally performing modalities. 
This ensures that for the lowest-performing modality, where $s_{m;b}^{t}$ is minimal, $\Delta_m$ will be 0.

% Since at the beginning of each epoch $t$, directly computing $Tr(F_m^t)$ is nontrivial, as this requires the full set of gradients for the epoch. Instead, MMPrime approximates the evaluation of whether a modality is in the prime learning window by using historic information from last epoch $Tr(F_m^{t-1})$.

% Then, the value of $\alpha$ is defined by:
% \begin{equation}
% \label{equ:score_alpha}
% \alpha  = exp\left ( \beta \ast \tanh \left ( \Delta_m \right )  \right ), 
% \end{equation} where $\beta$ is a hyperparameter controlling the sensitivity of $\alpha$  to the performance gap $\Delta_m$. 

% A higher value of $\alpha$ results in stronger regularization, thereby tightly constraining the rate of information acquisition for information-sufficient modalities. This adaptive regulation ensures balance and prevents any single modality from dominating during the prime learning window.

% In MMPrime, all adaptations are performed during the prime learning window. At the beginning of each epoch $t$, directly computing $Tr(F_i^t)$ is nontrivial, as this requires the full set of gradients for the epoch. Instead, MMPrime approximates the evaluation of whether a modality is in the prime learning window by using $Tr(F_i^{t-1})$ from the previous epoch. This approximation provides a practical method for determining the prime learning window during training. 

After identifying information-sufficient modalities based on $s_{m;b}^t$, the following criterion is used to determine whether these modalities are in the prime learning window:
\begin{equation} \label{equ:prime} 
\frac{Tr(F_m^{t-1}) - Tr(F_m^{t-2})}{Tr(F_m^{t-1})} > K, 
\end{equation} where $K$ is a positive hyperparameter that controls the threshold for inclusion. 
Equation \ref{equ:prime} reflects the relative changing rate of $Tr(F_m)$. A large value of this rate indicates that the information amount in the unimodal encoder is rapidly increasing. Conversely, a small or negative value suggests that information acquisition has slowed down or is even decreasing.

\begin{figure}[t]
    \centering
    \includegraphics[width=1\linewidth]{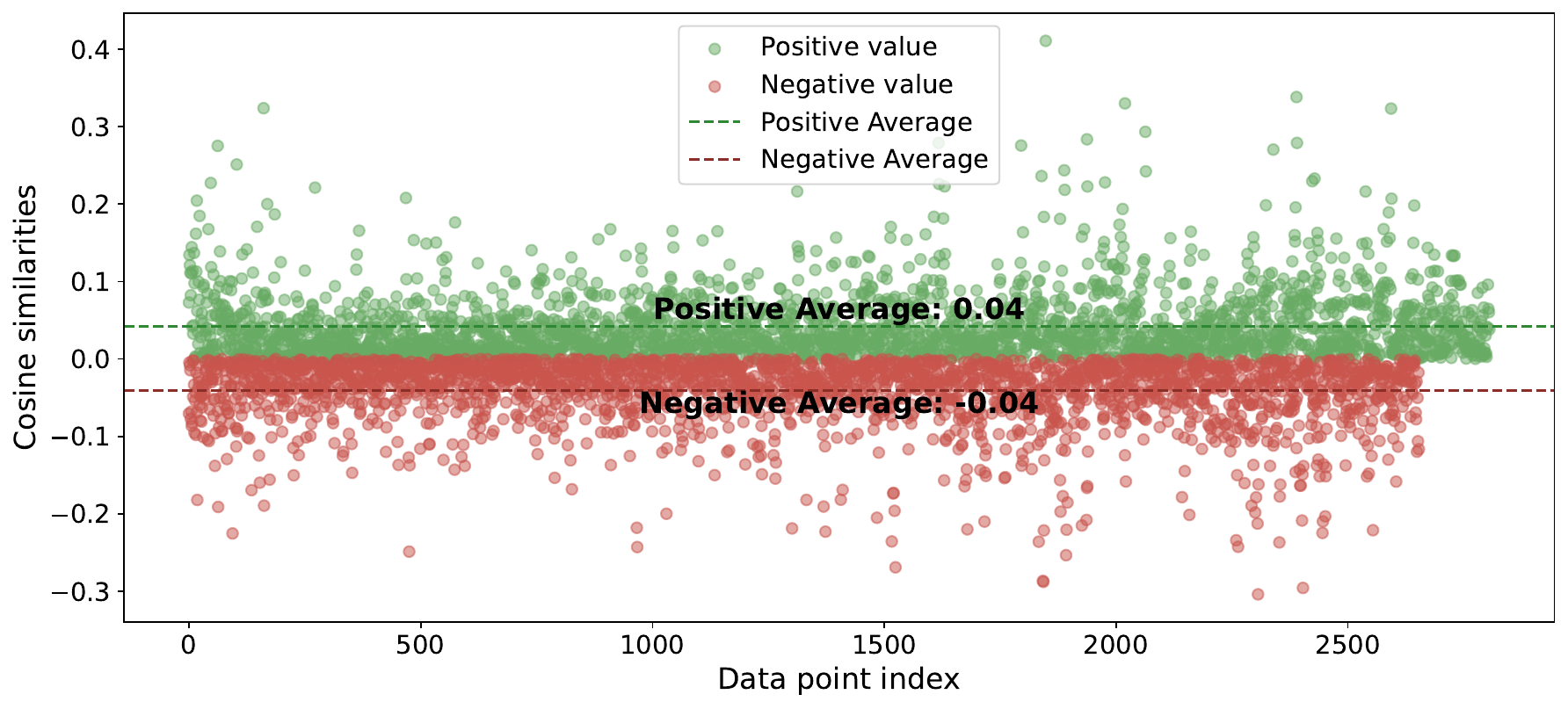}
    \vspace{-15pt}
    \caption{The cosine similarities of gradients across different batches within the prime learning window.}
    \label{fig:similarities}
\end{figure}

\begin{algorithm}[t]

    \renewcommand{\algorithmicrequire}{\textbf{Input:}}
    \renewcommand{\algorithmicensure}{\textbf{Output:}}
    \caption{Pipeline of InfoReg}
    \label{power}
    \begin{algorithmic}[1] % 控制是否有序号
        \REQUIRE Training dataset $D$, number of epochs $T$, number of batches $B$ of each batch, hyperparameter $\beta, K$
        % for loop

        \FOR {$t=0,1,\cdots,T-1$}
            \IF {$t <2$}
            \STATE Update model parameters;
            \STATE Calculate $Tr(F_m^t)$ by Equation \ref{equ:TrF};
            \STATE \textbf{Continue};
            \ENDIF
            \STATE Calculate $\frac{Tr(F_m^{t-1}) - Tr(F_m^{t-2})}{Tr(F_m^{t-1})}$;
            \FOR {$b=0,1,\cdots,B-1$}
                \STATE Randomly selects a batch of data from $D$;
                \STATE Calculate the performance scores $s_{m;b}^t$ for different modalities by Equation \ref{equ:score_j};
                \STATE Calculate $\Delta_m$ by Equation \ref{equ:delta};
                \STATE Decide information-sufficient modalities by $s_{m;b}^t$;
                \STATE Calculate $\alpha$ by Equation \ref{equ:score_alpha};
                \IF{$\frac{Tr(F_m^{t-1}) - Tr(F_m^{t-2})}{Tr(F_m^{t-1})} > K$ \textbf{and} $\Delta_m > 0$}
                    \STATE Calculate the regulation term by Equation \ref{equ:p};
                    \STATE Add adaptive unimodal regulation term;
                \ENDIF
                \STATE Update model parameters;
            \ENDFOR

        \STATE Calculate $Tr(F_m^t)$ by Equation \ref{equ:TrF}.
        \ENDFOR
        
    \end{algorithmic}
    \label{alg:algorithm1}
\end{algorithm}

\subsubsection{Adaptive unimodal regulation.}
\label{sec:adaptive unimodal regulation}

During the prime learning window, information-sufficient modalities dominate the model’s information acquisition, limiting the ability of information-insufficient modalities to acquire information. Therefore, it becomes essential to regulate information acquisition of information-sufficient modalities in this window.  However, directly calculating $Tr(F_m^t)$ requires complete gradient information across the entire dataset, making it impractical to regulate effectively. Additionally, calculating $Tr(F_m^t)$ over a full epoch is ineffective for adjusting the model at each iteration. To address this, we introduce a regulation term $P_{m;b}^t$ to approximately regulate the Fisher Information:
\begin{equation}
\label{equ:p}
    P_{m;b}^t = \frac{\alpha}{2} \parallel w_{m;b}^{t} - w_m^{t-1} \parallel ^2,
\end{equation} where $\alpha$ is a parameter that controls the degree of regulation.
This regulation term is proportional to $Tr(F_{m;b}^t)$ for each batch. Let the gradient $g_{\varphi^{m}}$ be denoted as $g$, $Tr(F_{m;b}^t)$ in each batch can be defined as: 
\begin{equation}
    Tr(F_{m;b}^t) = \frac{1}{b} \sum_{k=0}^{b} \parallel  g_k^t  \parallel ^2.
\label{equ:trf_g}
\end{equation} The regulation term $P_{m;b}^t$ can be written as:
\begin{align}
\label{equ:p2}
P_{m;b}^t &= \frac{\alpha}{2} \parallel w_{m;b}^{t} - w_m^{t-1} \parallel ^2 \notag \\
      &= \frac{\alpha}{2} \parallel - \eta \sum_{k=0}^{b} g_k^t \parallel ^2 \notag \\
      &= \frac{\alpha \eta^2}{2} \left(\sum_{k=0}^{b} \parallel g_k^t \parallel ^2 + 2 \sum_{0 \le z < k \le b}g_z^t(g_k^t)^T \right),
\end{align} where $\eta$ represents the learning rate. The high dimensionality of $g_b^t$ results in the gradients $ g_z^t (g_k^t)^T$ becoming approximately orthogonal for any two batches $z$ and $k$. The detailed proof is provided in the Appendix \ref{sec:proof}. As illustrated in Figure \ref{fig:similarities}, we compare the cosine similarities between gradients of different batches, and the results confirm that they are approximately orthogonal.
Then, $P_{m;b}^t$ can be approximated as:
\begin{equation}
    P_{m;b}^t = \frac{\alpha \eta^2}{2}\sum_{k=0}^{b} \parallel g_k^t \parallel ^2.
\label{equ:p3}
\end{equation}
According to the Equation \ref{equ:trf_g} and Equation \ref{equ:p3}, the unimodal regulation term $P_{m;b}^t$ regulates ${Tr}(F_{m;b}^t)$, thereby limiting the amount of information acquired by the information-sufficient modalities.
To modulate the impact of $P_{m;b}^t$, we define $\alpha$ as a dynamic parameter: 
\begin{equation}
\label{equ:score_alpha}
\alpha  = exp\left ( \beta \ast \tanh \left ( \Delta_m \right )  \right ), 
\end{equation} where $\beta$ is a hyperparameter controlling the sensitivity of $\alpha$  to the performance gap $\Delta_m$. A higher value of $\alpha$ results in stronger regulation, thereby tightly constraining the information acquisition for information-sufficient modalities. This adaptive regulation ensures balance and prevents any single modality from dominating during the prime learning window. Additional analysis of the regulation term is provided in Appendix \ref{supp:gradient}. Overall, our method is shown in Algorithm \ref{alg:algorithm1} and illustrated in Figure \ref{fig:overview}.

\begin{table*}[t]
    \centering
    \captionsetup{justification=raggedright}
    \vspace{0.5em}
    \begin{tabular}{@{} c | c c c | c c c @{}}
        \bottomrule
        \multirow{2}{*}{\textbf{Method}} & 
        \multicolumn{3}{c|}{\textbf{CREMA-D}} & 
        \multicolumn{3}{c}{\textbf{Kinetics Sounds}} \\ 
       % \cmidrule(lr){2-4} \cmidrule(lr){5-7} 
         & \textbf{Accuracy} & \textbf{Acc audio} & \textbf{Acc video} & \textbf{Accuracy} & \textbf{Acc audio} & \textbf{Acc video} \\
        \hline
        Joint training      & 66.61 & 58.99 & 35.14 & 65.67 & 53.13 & 36.01 \\

        OGM \cite{peng2022balanced}                   & 68.70 & 56.84 & 39.52 & \underline{66.63} & 53.39 & \underline{40.16} \\
        Greedy \cite{wu2022characterizing}               & 67.82 & 59.17 & 40.17 & 66.54 & 53.15 & 37.82 \\
        PMR \cite{fan2023pmr}                  & 66.92 & 57.83 & 38.91 & 66.33 & \underline{53.42} & 36.17 \\
        AGM \cite{li2023boosting}                  & \underline{69.71} & \underline{59.32} & \underline{43.72} & 66.54 & 53.12 & 37.24 \\
        InfoReg        & \textbf{71.90} & \textbf{60.03} & \textbf{49.65} & \textbf{69.31} & \textbf{54.16} & \textbf{44.73} \\
        \toprule
    \end{tabular}
    \vspace{-5pt}
    \caption{\textbf{Comparison with imbalanced multimodal learning methods.} All the methods only use one multimodal loss.  Bold and underline represent the best and second best respectively. }
    \label{tab:acc_part1}
\vspace{-10pt}
\end{table*}

\begin{table*}[ht]
    \centering
    \captionsetup{justification=raggedright}
    \vspace{0.5em}
    \begin{tabular}{@{} c | c c c | c c c @{}}
        \bottomrule
        \multirow{2}{*}{\textbf{Method}} & 
        \multicolumn{3}{c|}{\textbf{CREMA-D}} & 
        \multicolumn{3}{c}{\textbf{Kinetics Sounds}} \\ 
        %\cmidrule(lr){2-4} \cmidrule(lr){5-7} 
         & \textbf{Accuracy} & \textbf{Acc audio} & \textbf{Acc video} & \textbf{Accuracy} & \textbf{Acc audio} & \textbf{Acc video} \\
        \hline
        Joint training      & 66.61 & 58.99 & 35.14 & 65.67 & 54.13 & 36.01 \\
        Joint training*        & 70.81 & 60.52 & 55.23 & 68.71 & 55.23 & 44.18 \\

        G-Blending \cite{wang2020makes}            & 69.11 & 60.14 & 51.29 & 68.33 & 54.22 & 42.31 \\
        
        MMPareto \cite{weimmpareto}             & \underline{73.08} & \underline{60.83} & \underline{58.92} & \underline{71.11} & \underline{56.47} &  \underline{53.39} \\
        InfoReg*         & \textbf{75.71} & \textbf{61.63} & \textbf{61.22} & \textbf{72.03} & \textbf{57.21} & \textbf{53.57} \\
        \toprule
    \end{tabular}
    \caption{\textbf{Comparison with imbalanced multimodal learning methods with unimodal loss.} \textbf{Joint training*} and \textbf{InfoReg*} denote Joint training with unimodal loss and InfoReg with unimodal loss, repectively. Bold and underline represent the best and second best.}
    \label{tab:acc_part2}
\vspace{-10pt}
\end{table*}

\section{Experiments}

\subsection{Dataset and experimental settings}
\textbf{CREMA-D} \cite{cao2014crema} is an emotion recognition dataset with recordings of actors expressing six emotions, providing audio-visual samples to examine how auditory and visual cues convey emotion.
\textbf{Kinetics Sounds} \cite{arandjelovic2017look,kay2017kinetics} is designed for human action recognition, featuring 31 action classes from varied video sources, allowing analysis of audio-visual integration in dynamic activity recognition.
\textbf{CMU-MOSI} \cite{zadeh1606mosi}  is a sentiment analysis dataset with short video clips including audio, visual, and text modalities, suitable for exploring multimodal sentiment expression.

\begin{figure*}[t]
\centering
\begin{subfigure}{0.3\textwidth}  % 将宽度设置为 0.33
    \centering
    \includegraphics[width=\linewidth]{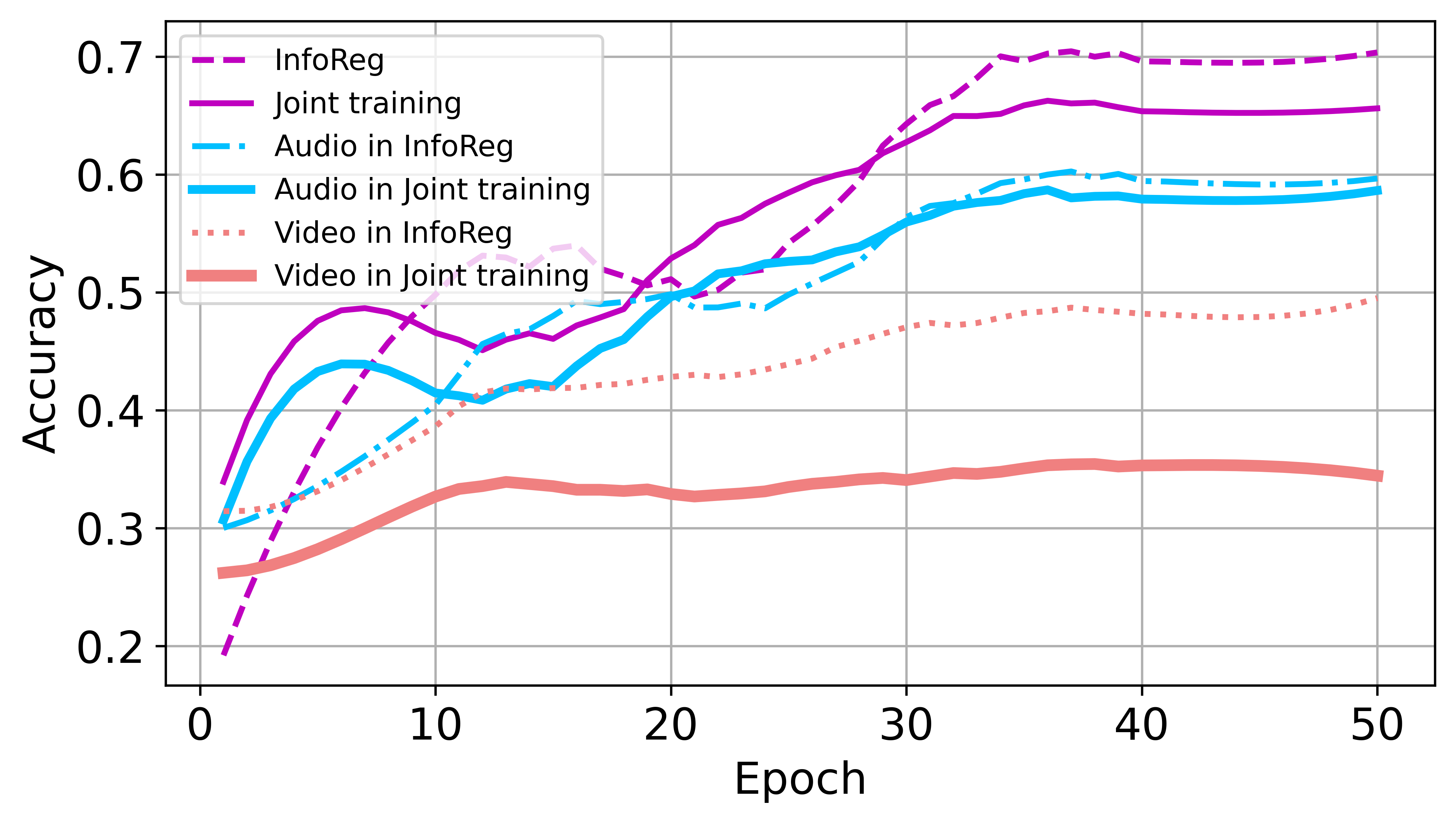}  % 替换为 accuracy 图片路径
    \vspace{-15pt}
    \caption{Comparison of accuracy.}
    \label{fig:accuracy}
\end{subfigure}%
\begin{subfigure}{0.3\textwidth}  % 将宽度设置为 0.33
    \centering
    \includegraphics[width=\linewidth]{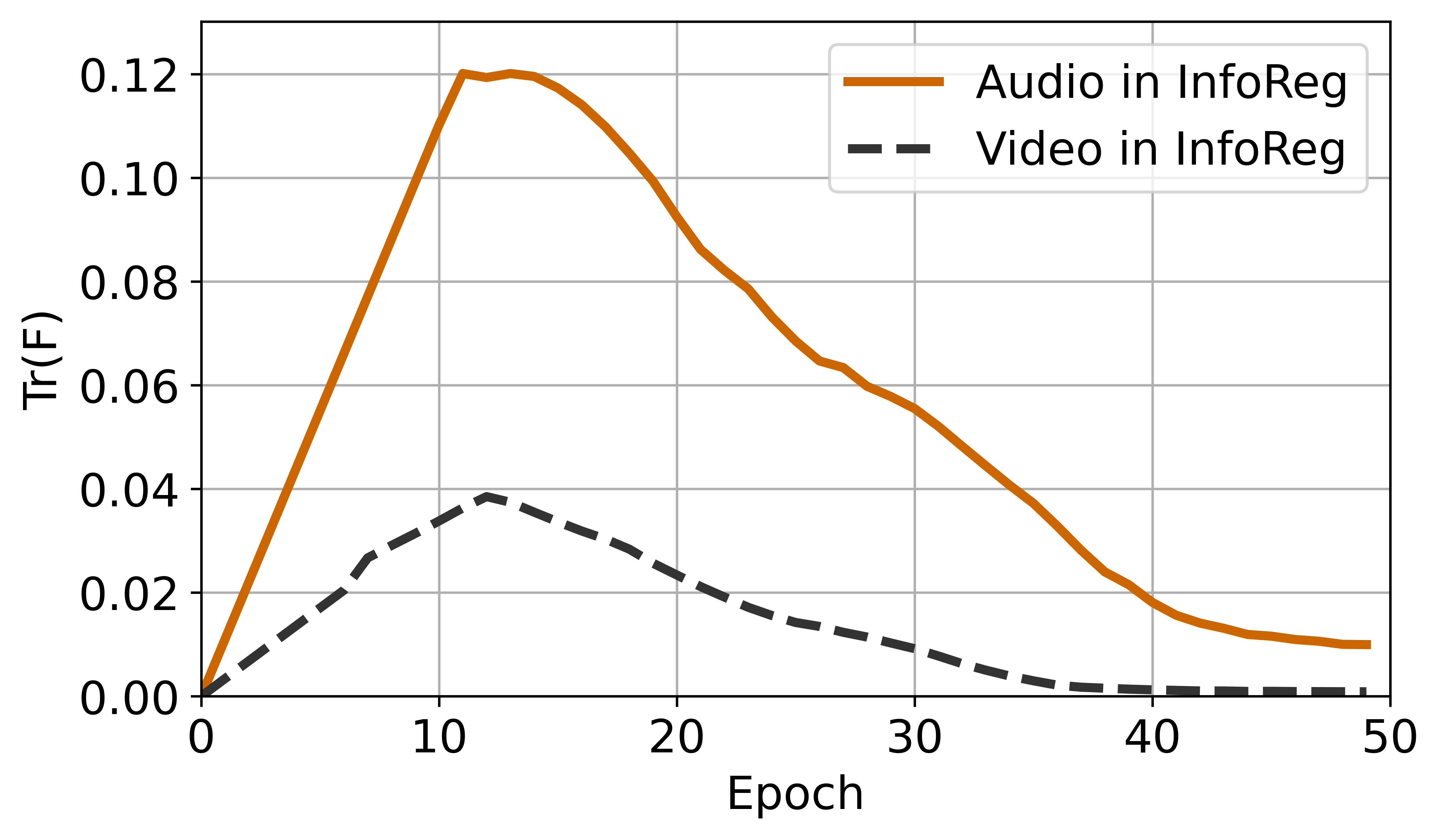}  % 替换为 mmclp3 图片路径
    \vspace{-15pt}
    \caption{$Tr(F)$ for both modalities.}
    \label{fig:mmclp3}
\end{subfigure}%
\begin{subfigure}{0.3\textwidth}  % 将宽度设置为 0.33
    \centering
    \includegraphics[width=\linewidth]{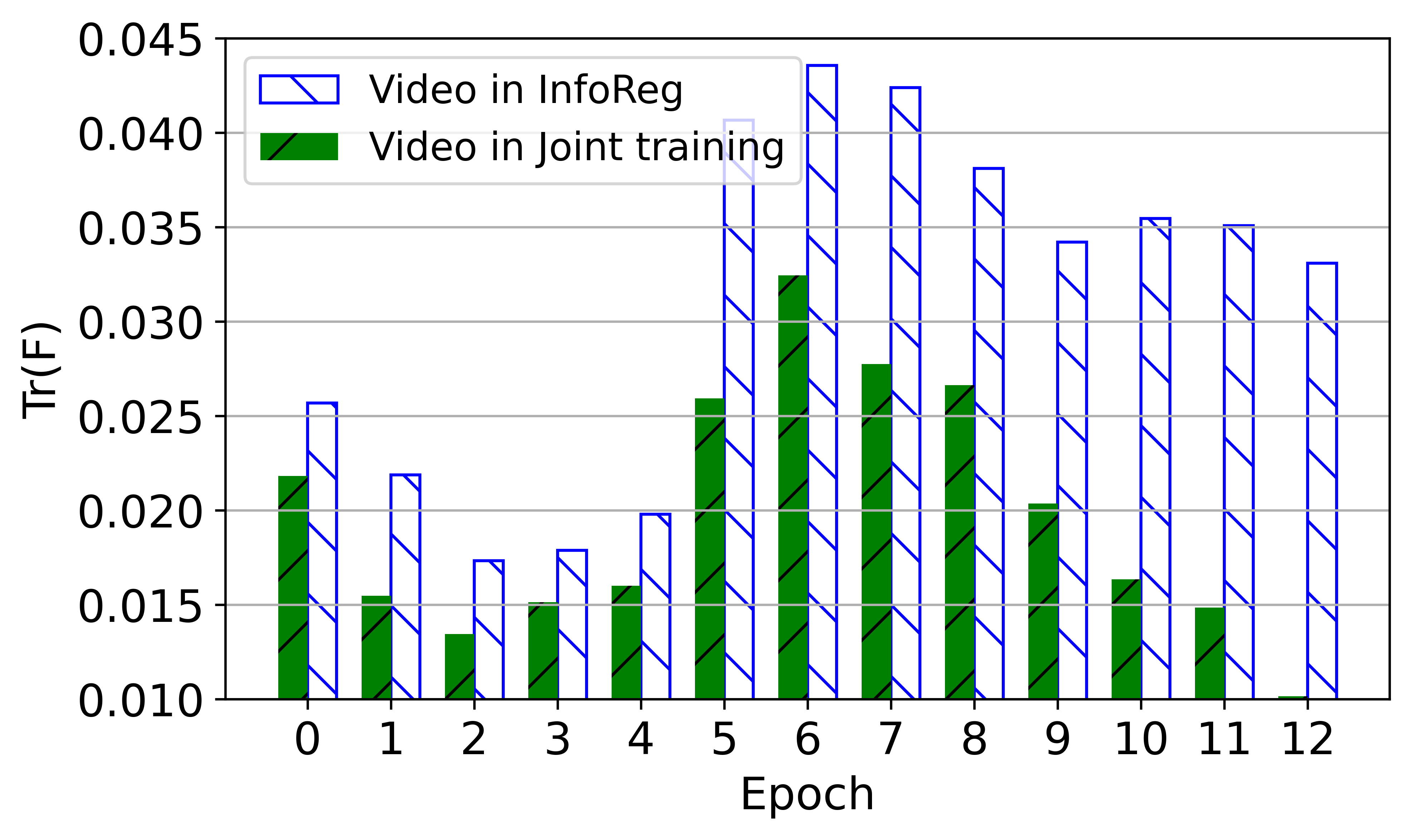}  % 替换为 TrF 图片路径
    \vspace{-15pt}
    \caption{Comparison of $Tr(F)$.}
    \label{fig:trf}
\end{subfigure}
\vspace{-8pt}
\caption{\textbf{(a).} The overall accuracy, audio accuracy, and video accuracy of InfoReg are compared with Joint training. \textbf{(b).} The value of $Tr(F)$ in InfoReg for both modalities. \textbf{(c).} The value of $Tr(F)$ of the video modality in InfoReg compared with that of Joint training. \textbf{All experiments are conducted on CREMA-D.}}
\vspace{-5pt}
\label{fig:comparison_all}
\end{figure*}

For our model architecture, we employ ResNet-18 \cite{he2016deep} as the backbone for the CREMA-D and Kinetics Sounds, while for the CMU-MOSI, we use a transformer-based model \cite{liang2021multibench}. All models are trained from scratch to ensure that the feature extraction processes are fully optimized for our specific tasks and datasets. Additionally, we implement a late fusion method to integrate uni-modal features from different modalities.

\subsection{Comparison with related imbalanced methods}
To evaluate the effectiveness of InfoReg in addressing information acquisition imbalance across modalities, we compare InfoReg with several imbalanced multimodal learning approaches, including G-Blending \cite{wang2020makes}, OGM \cite{peng2022balanced}, Greedy \cite{wu2022characterizing}, PMR \cite{fan2023pmr}, AGM \cite{li2023boosting}, and MMPareto \cite{weimmpareto}. Of these methods, G-Blending \cite{wang2020makes} and MMPareto \cite{weimmpareto} utilize both unimodal and multimodal losses, while OGM \cite{peng2022balanced}, Greedy \cite{wu2022characterizing}, PMR \cite{fan2023pmr}, and AGM \cite{li2023boosting} rely only on unimodal loss.
In our evaluation framework, \textbf{Joint training} denotes the widely used baseline for imbalanced multimodal learning, utilizing concatenation fusion with a single multimodal cross-entropy loss function \cite{peng2022balanced, li2023boosting}. Meanwhile, \textbf{Joint training*} represents the scenario where both unimodal and multimodal joint losses are applied simultaneously.

As shown in Table \ref{tab:acc_part1} and Figure \ref{fig:accuracy}, we conduct experiments on CREMA-D and Kinetics Sounds using InfoReg and several related methods that employ only multimodal loss. Based on the results, we first observe that all imbalanced methods achieve improved performance, indicating the significance of the modality imbalance issue and the need for balancing unimodal learning. Furthermore, InfoReg consistently outperforms other methods, especially with a notable improvement in the video modality. This suggests that regulating information acquisition during the prime learning window effectively balances information gain across modalities, enabling information-sufficient modalities to acquire more information and enhancing overall model performance.

\begin{figure*}[ht]
\centering
\begin{subfigure}{0.3\textwidth}  % 将宽度设置为 0.24
    \centering
    \includegraphics[width=\linewidth]{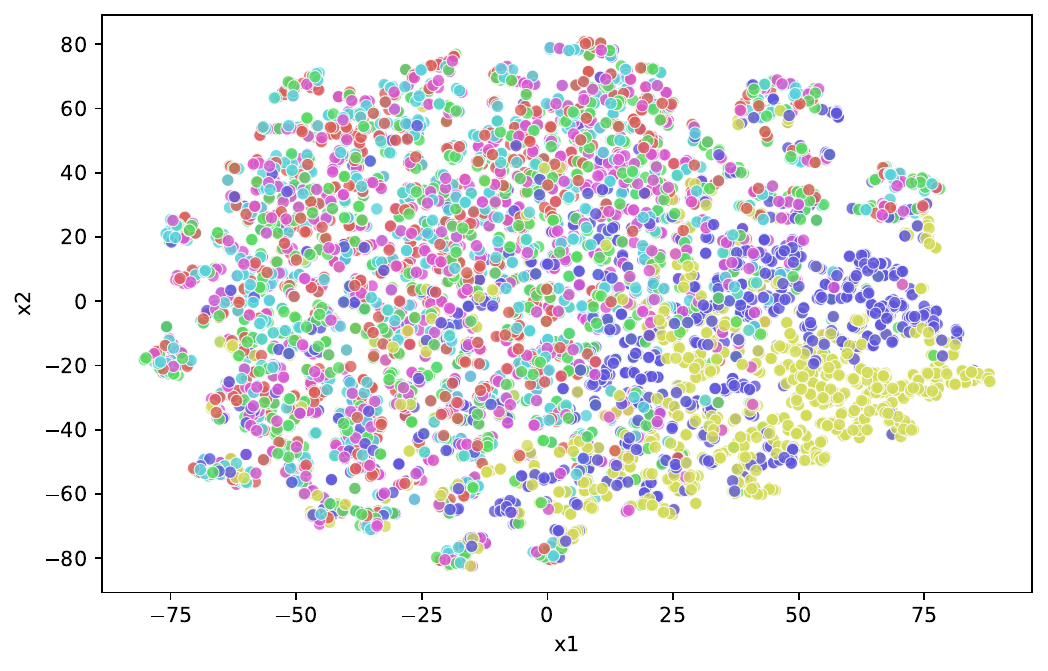}  % 替换为 TrF 图片路径
    \vspace{-15pt}
    \caption{InfoReg.}
    \label{fig:tsne1}
\end{subfigure}%
\begin{subfigure}{0.3\textwidth}  % 将宽度设置为 0.24
    \centering
    \includegraphics[width=\linewidth]{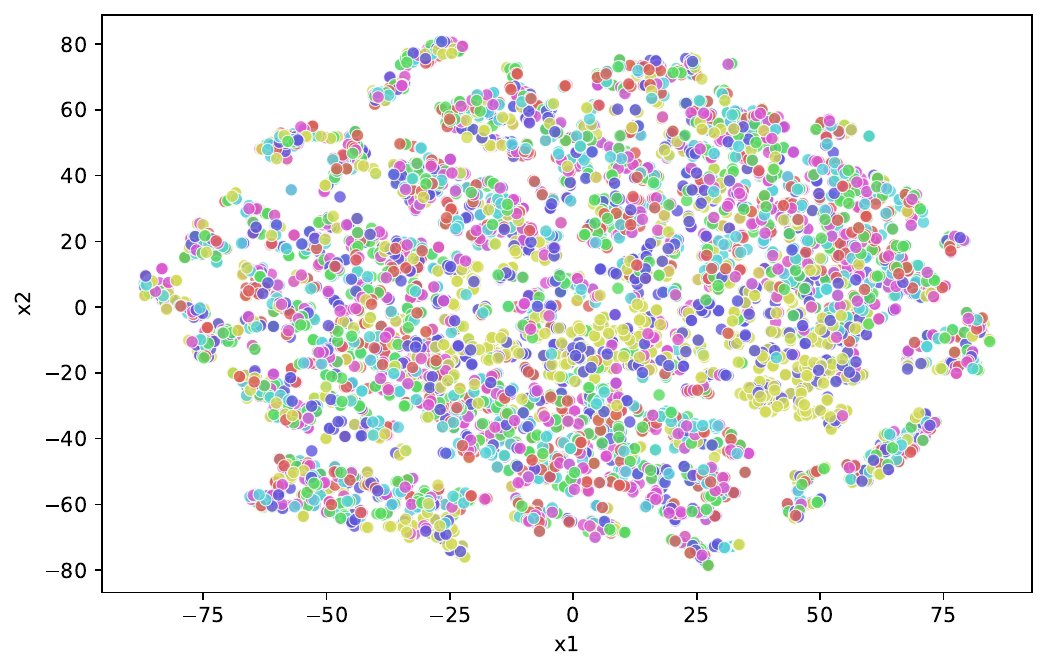}  % 替换为 mmclp3 图片路径
    \vspace{-15pt}
    \caption{Joint training.}
    \label{fig:tsne2}
\end{subfigure}%
\begin{subfigure}{0.3\textwidth}  % 将宽度设置为 0.24
    \centering 
    \includegraphics[width=\linewidth]{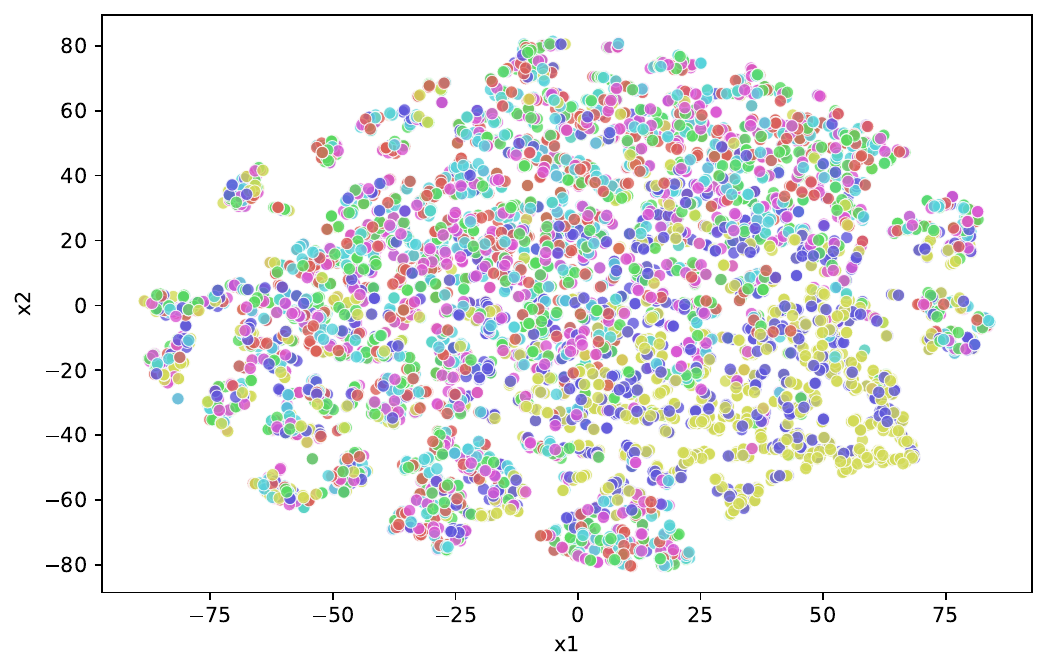}  % 替换为 your_new_image 图片路径
    \vspace{-15pt}
    \caption{Extended Joint training.}
    \label{fig:tsne3}
\end{subfigure}%
\vspace{-8pt}
\caption{The representations of the video modality on CREMA-D by t-SNE \cite{van2008visualizing} in InfoReg and Joint training are shown. "Extended Joint training" refers to Joint training that is extended to 100 epochs. Additional t-SNE representations are provided in Appendix \ref{supp:tsne}.}
\label{fig:tsne}
\vspace{-10pt}
\end{figure*}

Additionally, we provide experiments with methods  incorporating both multimodal loss and unimodal loss. Here, \textbf{Joint training*} and \textbf{InfoReg*} indicate both multimodal loss and unimodal loss are applied. As shown in Table \ref{tab:acc_part2}, all methods with unimodal loss achieve notable improvements over Joint training. This improvement is attributed to the inclusion of unimodal loss, which facilitates more efficient information retrieval from individual modalities, thereby enhancing overall performance. Moreover, InfoReg* continues to outperform other approaches and achieve better modality balance. This indicates that, with unimodal assistance, InfoReg can still effectively balance information acquisition across modalities, thereby achieving good modality balance and model performance.

% \begin{table}[t]
% \centering
% \begin{tabular}{c c c c}
% \toprule
% \multirow{2}{*}{\textbf{Method}} & \multicolumn{3}{c}{\textbf{CREMA-D}} \\
% \cmidrule(lr){2-4}
%  & \textbf{Accuracy} & \textbf{Acc audio} & \textbf{Acc video} \\
% \midrule
% Vanilla baseline & 70.81 & 60.52 & 55.23 \\
% OGM \cite{peng2022balanced} & 70.92 & 61.32 & 55.83 \\
% PMR \cite{fan2023pmr} & 72.61 & 60.33 & 58.12 \\
% AGM \cite{li2023boosting} & 71.31 & 60.87 & 56.97 \\
% InfoReg & \textbf{75.71} & \textbf{61.63} & \textbf{61.22} \\
% \bottomrule
% \end{tabular}
% \caption{\textbf{Comparison with imbalanced multimodal learning methods with extra uni-modal loss.} Bold values indicate the best results.}
% \label{tab:extended}
% \end{table}

% To ensure a fair comparison between InfoReg and other state-of-the-art approaches, we added uni-modal loss to several of these methods while keeping other settings unchanged. Under this configuration, all methods show an improvement in performance, as indicated in Table \ref{tab:extended}. Nevertheless, our method still outperforms all other state-of-the-art approaches and demonstrates better modality balance.

\subsection{Evaluating information acquisition}
We conduct experiments on CREMA-D to evaluate the effectiveness of InfoReg in enhancing information acquisition for information-insufficient modalities during the prime learning window. \textbf{Firstly}, As shown in Figure \ref{fig:mmclp3}, we observe that both modalities exhibit gradually growing values of $Tr(F)$ within the prime learning window, indicating that each modality acquires sufficient information early on. On one hand, the information-sufficient modality, audio, continues to acquire adequate information despite the regulation during the prime learning window, allowing it to maintain good performance throughout training. On the other hand, the information-insufficient modality also acquires sufficient information, as InfoReg alleviates the suppression of its information acquisition by the information-sufficient modality.
\textbf{Secondly}, we compare the $Tr(F)$ values for the video modality in InfoReg during the prime learning window with those in Joint training. As shown in Figure \ref{fig:trf}, the $Tr(F)$ values for the video modality in InfoReg are consistently higher than those in Joint training, indicating a considerable improvement in information acquisition for the video modality during the prime learning window. This enhancement enables the multimodal model to learn more comprehensive information during the prime learning window, thereby improving the model's overall performance.

\begin{table}[t]
\centering
\begin{tabular}{c|c c c}
\bottomrule
\multirow{2}{*}{\textbf{Method}} & \multicolumn{3}{c}{\textbf{CREMA-D}} \\
%\cmidrule(lr){2-4}
 & \textbf{WTP} & \textbf{PLW} & \textbf{Other periods} \\
\hline
Joint training & 66.61 & - & - \\
OGM \cite{peng2022balanced} & 68.70 & 68.76 & 67.13 \\
AGM \cite{li2023boosting} & \textbf{69.71} & \underline{69.54} & \textbf{67.41} \\
PMR \cite{fan2023pmr} & 66.92 & 66.99 & 66.57 \\
InfoReg & \underline{69.03} & \textbf{71.90} & \underline{67.22} \\
\toprule
\end{tabular}
\vspace{-5pt}
\caption{\textbf{Comparison of performance.} "WTP" and "PLW" denotes the whole training process and the prime learning window respectively. "Other periods" indicates adjustments made outside the prime learning window. Bold and underline represent the best and second best.}
\label{tab:prime}
\vspace{-10pt}
\end{table}

\begin{table}[t]
\centering
\begin{tabular}{c|cccc}
\bottomrule
\multirow{2}{*}{\textbf{Method}} & \multicolumn{4}{c}{\textbf{Fusion strategies}} \\
 & \textbf{Gated} & \textbf{SUM} & \textbf{FiLM} & \textbf{Concat} \\
\hline
Joint training & 65.32 & 64.38 & 66.67 & 66.61 \\
InfoReg & 69.18 & 70.12 & 70.23 & 71.90 \\
\toprule
\end{tabular}
\vspace{-5pt}
\caption{\textbf{Comparison of different fusion strategies.}}
\label{tab:fusion}
\vspace{-15pt}
\end{table}

\begin{table}[t]
\centering
\begin{tabular}{c|c c}
\bottomrule
\multirow{2}{*}{\textbf{Method}} & \multicolumn{2}{c}{\textbf{CMU-MOSI}} \\
%\cmidrule(lr){2-3}
 & \textbf{Accuracy} & \textbf{Macro F1} \\
\hline
Joint training & 61.09 & 60.74 \\
OGM \cite{peng2022balanced} & \underline{61.88} & \underline{61.32} \\
PMR \cite{fan2023pmr} & 61.47 & 60.98 \\
AGM \cite{li2023boosting} & 61.39 & 60.43 \\
InfoReg & \textbf{62.31} & \textbf{62.03} \\
\toprule
\end{tabular}
\vspace{-8pt}
\caption{\textbf{Comparison with imbalanced multimodal learning methods on the CMU-MOSI dataset.} Bold and underline represent the best and second best respectively.}
\label{tab:cmu-mosi}
\vspace{-10pt}
\end{table}

\begin{figure}[t]
\centering
\begin{subfigure}{0.48\columnwidth}  % 将宽度调整为 0.48
    \centering
    \includegraphics[width=\linewidth]{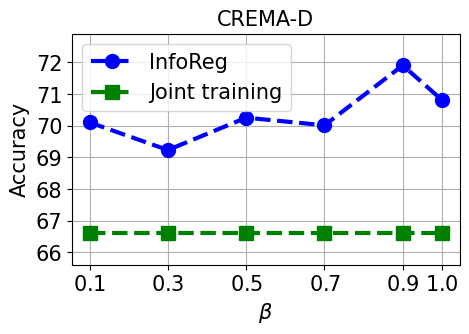}  % 替换为 mmclp2 图片路径
    \vspace{-15pt}
    \caption{Accuracy with different $\beta$.}
    \label{fig:beta cremad}
\end{subfigure}
\hspace{0.01\columnwidth}  % 调整间距
\begin{subfigure}{0.48\columnwidth}  % 将宽度调整为 0.48
    \centering
    \includegraphics[width=\linewidth]{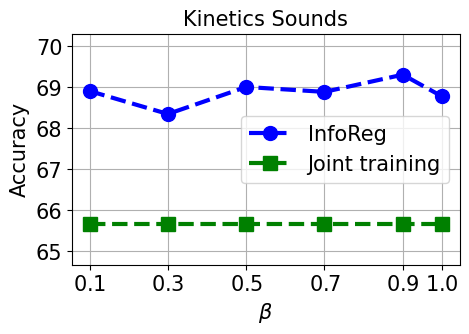}  % 替换为 TrF 图片路径
    \vspace{-15pt}
    \caption{Accuracy with different $\beta$.}
    \label{fig:beta ks}
\end{subfigure}
\vspace{-8pt}
\caption{\textbf{(a).} The overall accuracy of different $\beta$ on CREMA-D. \textbf{(b).} The overall accuracy of different $\beta$ on Kinetics Sounds.}
\vspace{-10pt}
\label{fig:ablation1}
\end{figure}

\begin{figure}[ht]
\centering
\begin{subfigure}{0.48\columnwidth}  % 将宽度调整为 0.48
    \centering
    \includegraphics[width=\linewidth]{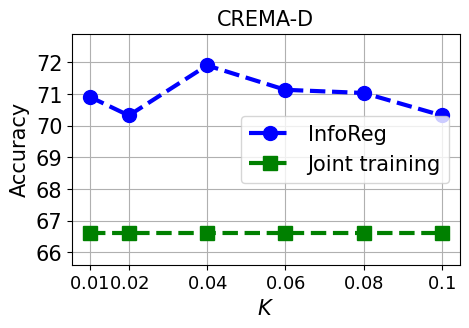}  % 替换为 mmclp2 图片路径
    \vspace{-15pt}
    \caption{Accuracy with different $K$.}
    \label{fig:K cremad}
\end{subfigure}
\hspace{0.01\columnwidth}  % 调整间距
\begin{subfigure}{0.48\columnwidth}  % 将宽度调整为 0.48
    \centering
    \includegraphics[width=\linewidth]{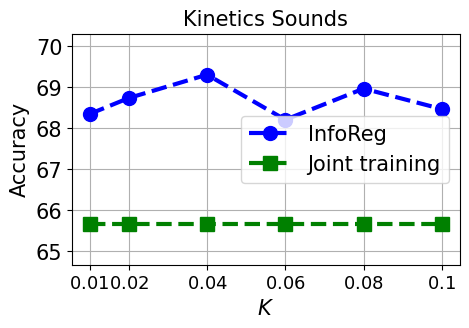}  % 替换为 TrF 图片路径
    \vspace{-15pt}
    \caption{Accuracy with different $K$.}
    \label{fig:K ks}
\end{subfigure}
\vspace{-8pt}
\caption{\textbf{(a).} The overall accuracy of different $K$ on CREMA-D. \textbf{(b).} The overall accuracy of different $K$ on Kinetics Sounds.}
\vspace{-18pt}
\label{fig:ablation2}
\end{figure}

\subsection{Importance of the prime learning window}
\label{sec:prime}
Unlike other methods, InfoReg introduces adjustments exclusively during the prime learning window. To evaluate the importance of the prime learning window in addressing the imbalanced multimodal learning problem, we compare the performance of several related methods with adjustments made exclusively during the prime learning window and during other periods. \textbf{Firstly}, all methods show considerable improvements when adjustments are made exclusively during the prime learning window. Specifically, the performance achieved by these methods during the prime learning window is comparable to that achieved by adjusting throughout the whole training process. This highlights the necessity of making adjustments within the prime learning window.
\textbf{Secondly}, adjustments made during other periods show significantly lower performance compared to those made during the prime learning window and the whole training process, indicating that late-stage adjustments are unnecessary due to the lack of additional information at this stage. Overall, our method focuses on balancing information acquisition exclusively within the prime learning window, effectively enhancing underrepresented modalities during this period and resulting in strong performance.

To further evaluate the importance of the prime learning window, we  use t-SNE \cite{van2008visualizing} to compare the features learned by the video modality using InfoReg on CREMA-D against those learned through Joint training and extended Joint training. As illustrated in Figure \ref{fig:tsne}, InfoReg yields higher-quality features due to sufficient information acquisition in the prime learning window.This aligns with previous studies, which have concluded that the quality of feature learning is highly correlated with the amount of information acquired early in training \cite{achille2018emergence}. Notably, as shown in Figure \ref{fig:tsne2} and Figure \ref{fig:tsne3}, extending the training period does not compensate for the information loss experienced by the video modality during the prime learning window. Our experiments highlight the critical role of the prime learning window in information acquisition.

\vspace{-3pt}

\subsection{The influence of fusion strategies}

\vspace{-3pt}
We evaluate the performance of InfoReg with four different fusion strategies, including Gated \cite{kiela2018efficient}, SUM, FiLM \cite{perez2018film}, and Concat. For the Gated, SUM, and FiLM strategies, we measured the performance of each modality individually by zeroing out the features of other modalities. As shown in Table \ref{tab:fusion}, InfoReg demonstrates consistently robust performance across various fusion strategies. This highlights the adaptability and scalability of InfoReg.

\vspace{-3pt}

\subsection{Extension to more complex settings}
\vspace{-3pt}
To validate that our method continues to perform well in more complex settings, we conducted experiments on CMU-MOSI using a Transformer architecture following \cite{liang2021multibench}. As shown in Table \ref{tab:cmu-mosi}, InfoReg continues to perform effectively in these more challenging scenarios, outperforming other related methods. This demonstrates the good scalability of InfoReg, highlighting its ability to extend to more complex transformer-based architectures and scenarios involving more than two modalities.
Further experiments exploring dominant modalities and scenarios requiring inter-modality cooperation are provided in Appendix \ref{supp:exp}.

\vspace{-3pt}

\subsection{Hyperparameter sensitivity analysis}

% The hyperparameter $\beta$ controls the strength of the proximal term, chosen from candidate values $\{0.1, 0.3, 0.5, 0.9, 1.0\}$. A larger $\beta$ increases the proximal term, intensifying suppression of the information-sufficient modality. If $\beta$ is too small, early-stage suppression may be insufficient, leading to inadequate learning by the harder-to-train modality. Conversely, an excessively large $\beta$ can overly suppress the information-sufficient modality, limiting its ability to gather information during the prime learning window. Based on our experimental results shown in Figure \ref{fig:ablation1}, we selected $\beta = 0.9$, which provided the best accuracy while maintaining minimal accuracy fluctuations across different values. This highlights the robustness and effectiveness of our approach, consistently outperforming joint training.

% The hyperparameter $K$ determines the duration of the prime learning window, assessed using the slope of the trace of the Fisher Information Matrix. We identified the CLP as the period when the trace shows an upward trend, selecting $K$ from $\{0.01, 0.02, 0.04, 0.05, 0.1\}$. In our experiments, the prime learning window typically represented 15\% to 20\% of the total training time. Figure \ref{fig:ablation2} indicates that $K = 0.04$ was optimal. VaIARtions in $K$ also showed minimal impact on model accuracy, demonstrating the consistency and reliability of our method in outperforming state-of-the-art techniques.
\vspace{-3pt}

The hyperparameter $\beta$, controlling the regulation term's strength, is selected from $\{0.1, 0.3, 0.5, 0.7, 0.9, 1.0\}$. A larger $\beta$ increases the regulation degree of the information-sufficient modalities.  Our results (Figure \ref{fig:ablation1}) show that $\beta = 0.9$ provided the best accuracy. Similarly, the hyperparameter $K$, which serves as the threshold for determining whether the model is within the prime learning window, is chosen from $\{0.01, 0.02, 0.04, 0.06, 0.08, 0.1\}$. Figure \ref{fig:ablation2} indicates that $K = 0.04$ yields the best performance.

\vspace{-4pt}

\vspace{-3pt}
\section{Conclusion}
\vspace{-3pt}
% In this paper, we identify that there is a prime learning window in multimodal learning where one modality's information acquisition may be suppressed by others. To address this, we propose the Information Acquisition Regulation algorithm, which balances information acquisition across modalities by regulating the acquisition rates of information-sufficient modalities during this stage. Our method promotes a more balanced learning process, accordingly enhancing model performance. Experiments across multiple datasets show that our method alleviates imbalanced multimodal learning and achieves superior performance.
In this paper, we identify that there is a prime learning window in multimodal learning, and one modality's information acquisition can be suppressed by others during this stage. Then, we propose the Information Acquisition Regulation algorithm. It aims to address the imbalance in information acquisition across modalities by regulating the acquisition rates of information-sufficient modalities during the prime learning window. Our method promotes a more balanced learning process, accordingly enhancing model performance. Experiments across multiple datasets show that our method alleviates imbalanced multimodal learning and then achieves superior performance.

\section{Acknowledgment}
This work was supported through the National Natural Science Foundation of China (Grant No.62106272) and benefited from the research grant program of the CCF-Zhipu.AI Large Model Innovation Fund.
{
    \small
    \bibliographystyle{ieeenat_fullname}
    \bibliography{main}
}
\clearpage
\setcounter{page}{1}
\maketitlesupplementary

\appendix
% 
% 定理环境
\newtheorem{theorem}{Theorem}

\section{Orthogonality proof}
\label{sec:proof}

This proof supports the analysis presented in Section \ref{sec:adaptive unimodal regulation}, specifically Equation \ref{equ:p2}, where the regulation term \( P_{m;b}^t \) involves gradients  $\mathbf{g}_k^t$ from multiple batches. The analysis assumes that, due to the high dimensionality of the space, the gradients  $\mathbf{g}_k^t$ from different batches are nearly orthogonal. Here, we formally prove this assumption by showing that random vectors sampled from the surface of a high-dimensional hypersphere are nearly orthogonal with high probability.

\paragraph{Lemma 1.}
In high-dimensional spaces, let \( \mathbf{g}_z^t, \mathbf{g}_k^t \in \mathbb{R}^n \) be two random vectors uniformly sampled from the surface of an \( n \)-dimensional hypersphere with magnitudes \( \|\mathbf{g}_z^t\| = a \) and \( \|\mathbf{g}_k^t\| = b \). As \( n \to \infty \), these vectors are nearly orthogonal with high probability. Specifically, their dot product satisfies:
\begin{equation}
\mathbf{g}_z^t \cdot \mathbf{g}_k^t = ab \cos \theta \approx 0.
\end{equation}

\paragraph{Proof of Lemma 1.}
Let $\mathbf{g}_z^t$ and $\mathbf{g}_k^t$ be two random vectors in $\mathbb{R}^n$ with magnitudes $\|\mathbf{g}_z^t\| = a$ and $\|\mathbf{g}_k^t\| = b$. The dot product is given by:
\begin{equation}
\mathbf{g}_z^t \cdot \mathbf{g}_k^t = a b \cos\theta.
\end{equation}
To analyze the distribution of the angle $\theta$ in high-dimensional space, we consider the geometry of the $n$-dimensional unit hypersphere. Any vector $\mathbf{x} \in \mathbb{R}^n$ with unit norm, i.e., $\|\mathbf{x}\|_2 = 1$, lies on the surface of the unit hypersphere. It can be parameterized in spherical coordinates as:
\begin{equation}
\mathbf{x} = (x_1, x_2, \dots, x_n), \quad \text{where } x_i \in \mathbb{R}, \ \sum_{i=1}^n x_i^2 = 1.
\end{equation}
The components of $\mathbf{x}$ in spherical coordinates are:
\begin{equation}
\begin{aligned}
x_1 &= \cos\phi_1, \\
x_2 &= \sin\phi_1 \cos\phi_2, \\
x_3 &= \sin\phi_1 \sin\phi_2 \cos\phi_3, \\
&\ \vdots \\
x_n &= \prod_{i=1}^{n-1} \sin\phi_i,
\end{aligned}
\end{equation}
where $\phi_1, \phi_2, \dots, \phi_{n-2} \in [0, \pi]$, and $\phi_{n-1} \in [0, 2\pi]$. The surface element of the hypersphere is:
\begin{equation}
dS = (\sin\phi_1)^{n-2} (\sin\phi_2)^{n-3} \cdots \sin\phi_{n-2} \, d\phi_1 d\phi_2 \cdots d\phi_{n-1}.
\end{equation}
Without loss of generality, let one vector $\mathbf{g}_z^t$ be fixed along the $x_1$-axis, $\mathbf{g}_z^t = (a, 0, \dots, 0).$
The second vector $\mathbf{g}_k^t$ can be parameterized using spherical coordinates. The angle $\theta$ between $\mathbf{g}_z^t$ and $\mathbf{g}_k^t$ is the same as $\phi_1$, the first coordinate angle, so:
\begin{equation}
\cos\phi_1 = \cos\theta.
\end{equation}
The relevant term in the hypersphere surface element is:
\begin{equation}
p_n(\phi_1) \propto (\sin\phi_1)^{n-2}.
\end{equation}
This shows that the probability density of $\phi_1$ (or $\theta$) depends on the sine function raised to the power of $(n-2)$.
For large $n$, $(\sin\phi_1)^{n-2}$ is sharply concentrated around $\phi_1 = \pi/2$ because $\sin\phi_1$ reaches its maximum at $\pi/2$. As $n \to \infty$, this concentration becomes stronger, leading to $\phi_1 \approx \frac{\pi}{2}$ with high probability.
Since $\phi_1 \approx \pi/2$, we have:
\begin{equation}
\cos\phi_1 = \cos \theta\ \approx 0.
\end{equation}
Thus, in high-dimensional spaces, the angle $\theta$ between two random vectors concentrates around $\pi/2$, leading to:
\begin{equation}
\mathbf{g}_z^t \cdot \mathbf{g}_k^t 
 = a b\cos \theta\approx 0.
\end{equation}
This demonstrates that the vectors are nearly orthogonal as $n \to \infty$.

\section{Gradient norm analysis}
\label{supp:gradient}
This section aims to demonstrate that the regulation term \( P_{m;b}^t \), introduced to regulate the information-sufficient modalities during the prime learning window, does not hinder the convergence of the optimization process. Specifically, we analyze the gradient norm and show that, under proper parameter settings, the convergence rate remains consistent with that of the original optimization objective without the regulation term.

\paragraph{Lemma 2.}
At training epoch \( t \) and batch \( b \), consider the optimization objective:
\begin{equation}
\mathcal{L}(w_{m;b}^t) = \mathcal{L}_{joint}(w_{m;b}^t) + P_{m;b}^t,
\end{equation}
where \( \mathcal{L}_{joint}(w_{m;b}^t) \) is the multimodal joint loss function, and the regulation term \( P_{m;b}^t \) is defined as:
\begin{equation}
P_{m;b}^t = \frac{\alpha \eta^2}{2} \sum_{k=0}^b \| g_k^t \|^2.
\end{equation}
Here, \( \alpha > 0 \) is the regularization coefficient, \( \eta > 0 \) is the learning rate, and \( g_k^t \) denotes the gradient of batch \( k \) at epoch \( t \). If \( \alpha \) and \( \eta \) are sufficiently small, the convergence rate remains of the same order as without the regulation term.

\paragraph{Proof of Lemma 2.}
During the training, the weight update rule is given by:
\begin{equation}
w^{t}_{m;b+1} = w^t_{m;b} - \eta \nabla \mathcal{L}(w^t_{m;b}),
\end{equation}
where:
\begin{equation}
\nabla \mathcal{L}(w^t_{m;b}) = \nabla \mathcal{L}_{joint}(w^t_{m;b}) + \nabla P_{m;b}^t.
\end{equation}
The gradient of the regulation term \( P_{m;b}^t \) is given by:
\begin{equation}
\nabla P_{m;b}^t = \alpha \eta^2 \sum_{k=0}^b g_k^t.
\end{equation}
Assuming that \( \mathcal{L}(w) \) is \( L \)-Lipschitz smooth, we have:
\begin{equation}
\begin{aligned}
\mathcal{L}(w^{t}_{m;b+1}) \leq &\ \mathcal{L}(w^t_{m;b}) 
+ \nabla \mathcal{L}(w^t_{m;b})^T (w^{t}_{m;b+1} - w^t_{m;b}) \\
&+ \frac{L}{2} \| w^{t}_{m;b+1} - w^t_{m;b} \|^2.
\end{aligned}
\end{equation}
Substituting \( w^{t}_{m;b+1} - w^t_{m;b} = -\eta \nabla \mathcal{L}(w^t_{m;b}) \), we obtain:
\begin{equation}
\mathcal{L}(w^{t}_{m;b+1}) \leq \mathcal{L}(w^t_{m;b}) - \eta \|\nabla \mathcal{L}(w^t_{m;b})\|^2 + \frac{L \eta^2}{2} \|\nabla \mathcal{L}(w^t_{m;b})\|^2.
\end{equation}
The gradient norm is expressed as:
\begin{equation}
\begin{aligned}
\|\nabla \mathcal{L}(w^t_{m;b})\|^2 = &\ \|\nabla \mathcal{L}_{joint}(w^t_{m;b})\|^2 \\
&+ 2 \nabla P_{m;b}^t \cdot \nabla  \mathcal{L}_{joint}(w^t_{m;b}) \\
&+ \|\nabla P_{m;b}^t\|^2.
\end{aligned}
\end{equation}
Due to the high dimensionality of the space, as demonstrated in Section \ref{sec:proof}, the regulation term gradient \( \nabla P_{m;b}^t \) and the joint loss gradient \( \nabla \mathcal{L}_{joint}(w^t_{m;b}) \) are nearly orthogonal. As a result, their dot product can be approximated as:
\begin{equation}
\nabla P_{m;b}^t \cdot \nabla \mathcal{L}_{joint}(w^t_{m;b}) \approx 0.
\end{equation}
The gradient of the regulation term is bounded as:
\begin{equation}
\|\nabla P_{m;b}^t\| = \alpha \eta^2 \left\|\sum_{k=0}^b g_k^t\right\| \leq \alpha \eta^2 b G,
\end{equation}
where \( G \) is the upper bound of the gradient norm \( \| g_k^t \| \). Thus, the term satisfies:
\begin{equation}
\|\nabla \mathcal{L}(w^t_{m;b})\|^2 \leq \|\nabla \mathcal{L}_{joint}(w^t_{m;b})\|^2 + \alpha^2 \eta^4 b^2 G^2.
\end{equation}
For sufficiently small \( \alpha \) and \( \eta \), the additional term \( \alpha^2 \eta^4 b^2 G^2 \) becomes negligible. Therefore, the convergence rate remains of the same order as without \( P_{m;b}^t \).

\section{Supplementary t-SNE analysis}
\label{supp:tsne}
\begin{figure}[ht]
\centering
\begin{subfigure}{0.48\linewidth}  % 将宽度设置为 0.45
    \centering
    \includegraphics[width=\linewidth]{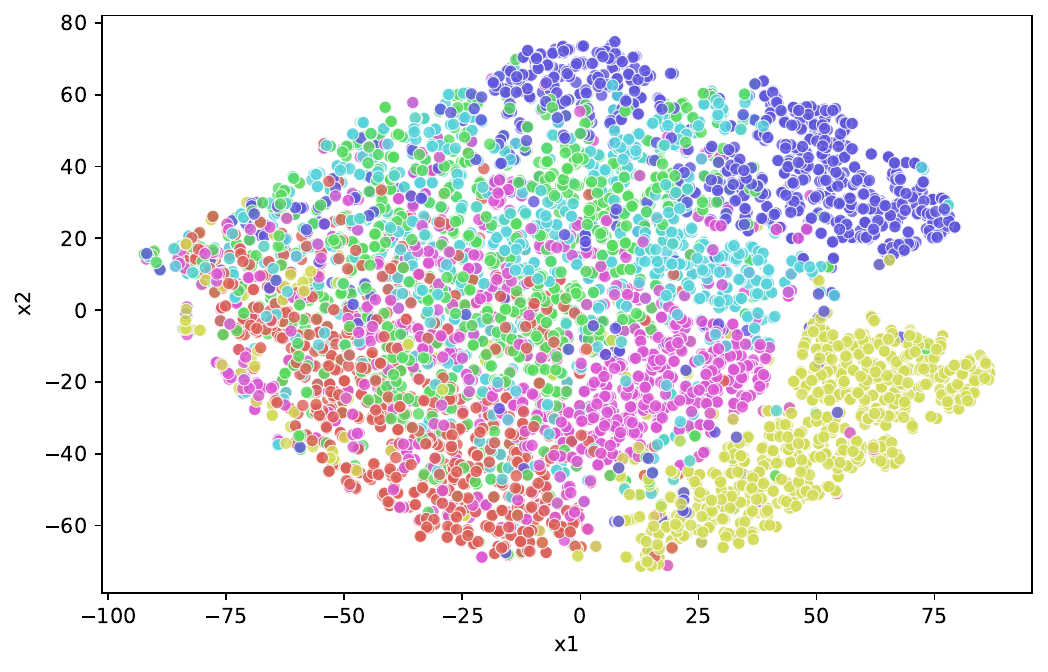}
    \caption{InfoReg*.}
    \label{fig:tsne*1}
\end{subfigure}%
\hspace{0.02\linewidth}  % 调整图片间的水平间距
\begin{subfigure}{0.48\linewidth}  % 将宽度设置为 0.45
    \centering
    \includegraphics[width=\linewidth]{video_unextended_tsne_rep.pdf}
    \caption{Joint training.}
    \label{fig:tsne*2}
\end{subfigure}

\vspace{0.3cm}  % 调整两行图片之间的垂直间距

\begin{subfigure}{0.48\linewidth}  % 将宽度设置为 0.45
    \centering
    \includegraphics[width=\linewidth]{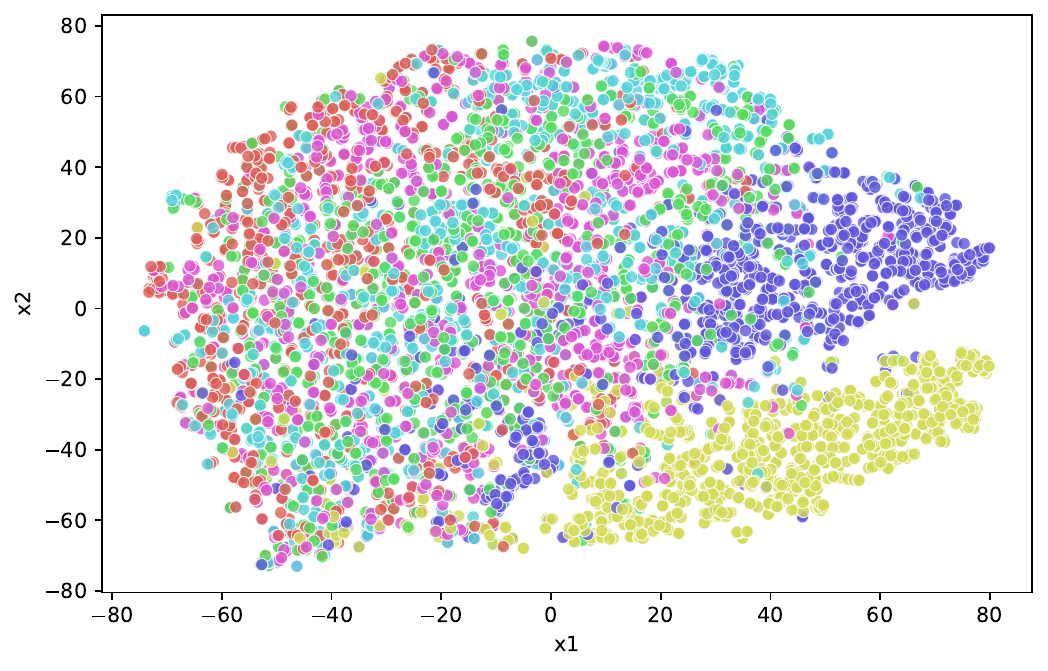}
    \caption{Joint training*.}
    \label{fig:tsne*3}
\end{subfigure}%
\hspace{0.02\linewidth}  % 调整图片间的水平间距
\begin{subfigure}{0.48\linewidth}  % 将宽度设置为 0.45
    \centering
    \includegraphics[width=\linewidth]{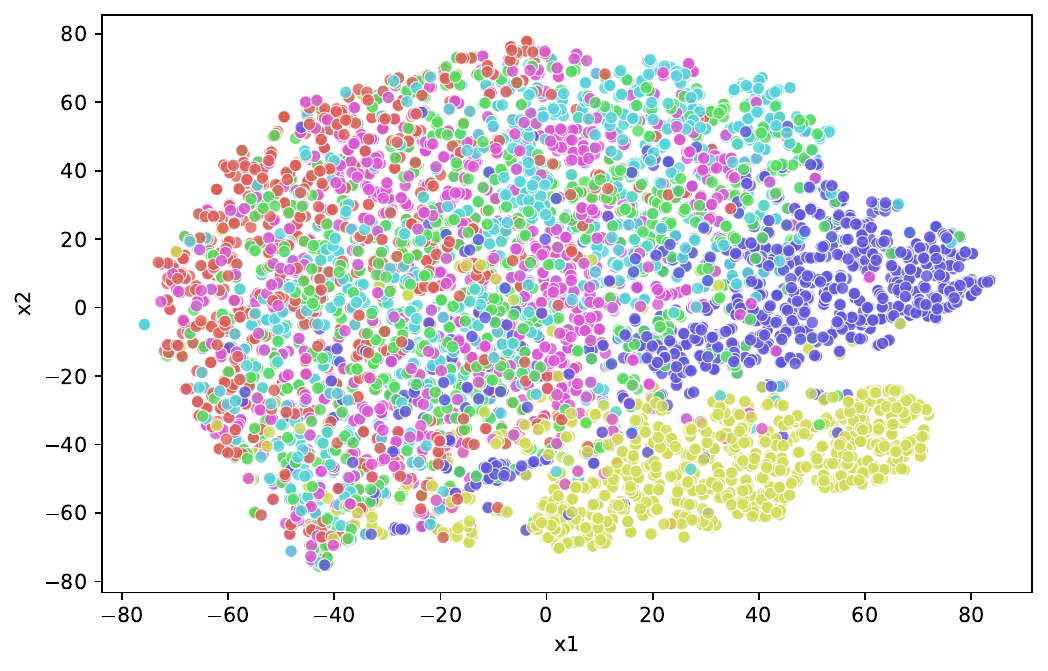}
    \caption{Extended Joint training*.}
    \label{fig:tsne*4}
\end{subfigure}

\caption{The representations of the video modality on CREMA-D by t-SNE \cite{van2008visualizing} across different methods are shown. InfoReg* and Joint training* denote InfoReg and Joint training with unimodal loss respectively. "Extended Joint training*" denotes Joint training* that is extended to 100 epochs.}
\label{fig:tsne*}
\end{figure}
To provide a more comprehensive evaluation of the proposed InfoReg method, we extend our analysis by incorporating t-SNE visualizations of video modality representations for InfoReg* and Joint training* on the CREMA-D dataset. Here, InfoReg* denotes InfoReg with unimodal loss, and Joint training* denotes Joint training with unimodal loss. As shown in Figure \ref{fig:tsne*}, InfoReg* and Joint training* learn better representations than Joint training. This is because the unimodal loss helps the multimodal model acquire more information. Additionally, the features learned by Joint training* and Extended Joint training* are similar, as shown in Figure \ref{fig:tsne*3} and Figure \ref{fig:tsne*4}. This indicates that extending the training time cannot compensate for the lack of information acquired during the prime learning window.  Furthermore, InfoReg* learns better representations than both Joint training* and Extended Joint training*.
This demonstrates that, with unimodal loss, our method can still help information-insufficient modalities acquire more information in the prime learning window. As a result, InfoReg* learns better representation.

\section{Supplementary experiments}

\begin{figure}[htbp] \centering \vspace{-12pt} \includegraphics[width=0.75\linewidth]{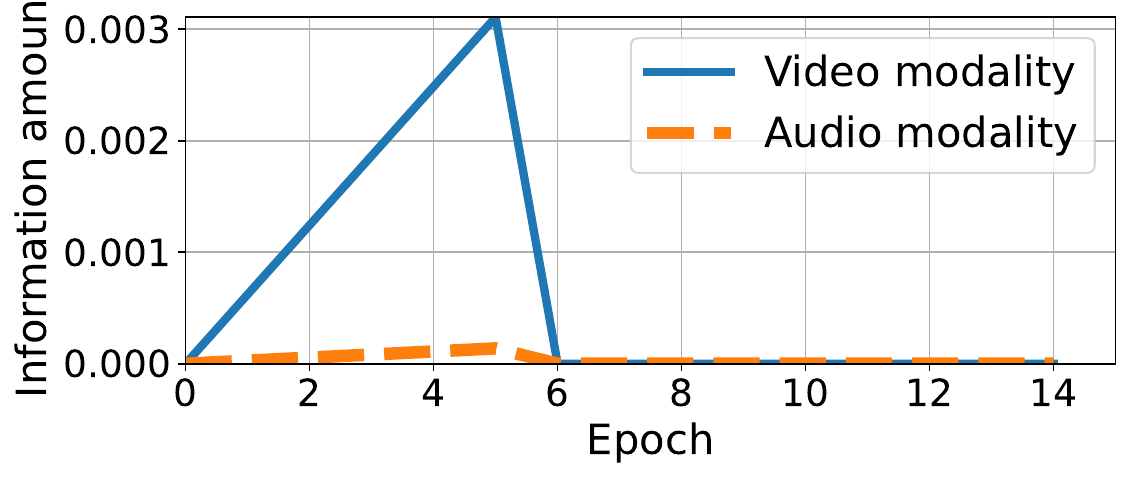} \vspace{-10pt} \caption{Violence Flow dataset example, showcasing video modality dominance.} \label{fig:mmclp-rebuttal} \end{figure}

\begin{table}[htbp] \centering \begin{tabular}{lcc} \hline \textbf{Dataset} & \textbf{Violence Flow} & \textbf{Hateful Memes} \\ \hline Joint training & 89.21 & 55.00 \\ InfoReg & \textbf{90.56} & \textbf{56.20} \\ \hline \end{tabular} \vspace{-8pt} \caption{Accuracy comparison.} \label{tab:hatememes} \end{table}

\label{supp:exp}
To further evaluate the effectiveness of InfoReg under diverse dataset conditions, we conducted experiments on the Violence Flow \cite{hassner2012violent} and Hateful Memes \cite{kiela2020hateful} datasets. These datasets present different challenges: Violence Flow emphasizes anomaly detection, where the video modality quickly becomes dominant, while Hateful Memes requires cooperation between modalities due to its complex multimodal nature.

Figure \ref{fig:mmclp-rebuttal} illustrates the information amount during training on the Violence Flow, where the video modality demonstrates dominance during the prime learning window. InfoReg can identify this dominant modality.

The Hateful Memes dataset requires significant cooperation between modalities. As shown in the Table \ref{tab:hatememes}, Despite the increased complexity, InfoReg can still improve the perforemance of the model.

% WARNING: do not forget to delete the supplementary pages from your submission 
% \input{sec/X_suppl}

\end{document}